\documentclass[11pt,oneside]{book}
\usepackage[margin=1.2in]{geometry}
\usepackage[toc,page]{appendix}
\usepackage{graphicx}
\usepackage{amsmath}
\usepackage{caption}
\usepackage{algorithm}
\usepackage{algpseudocode}
\usepackage{subcaption}
\usepackage{parskip}
\usepackage{amssymb}
\usepackage{xcolor}
\usepackage{hyperref}

\hypersetup{
    colorlinks=true,
    linkcolor=black,
    citecolor=black,
    urlcolor=black,
}

\newtheorem{definition}{Definition}
\newtheorem{theorem}{Theorem}

\begin{document}

\captionsetup[figure]{margin=1.5cm,font=small,labelfont={bf},name={Figure},labelsep=colon,textfont={it}}
\captionsetup[table]{margin=1.5cm,font=small,labelfont={bf},name={Table},labelsep=colon,textfont={it}}
% \setlipsumdefault{1}

\frontmatter

\begin{titlepage}

% -------------------------------------------------------------------
% You need to edit the details here
% -------------------------------------------------------------------

\begin{center}
{\LARGE University of Sheffield}\\[1.5cm]
\linespread{1.2}\huge {\bfseries Reinforcement Learning and Video Games}\\[1.5cm]
\linespread{1}
\includegraphics[width=5cm]{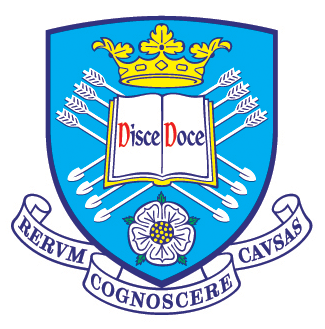}\\[1cm]
{\Large Yue Zheng}\\[1cm]
{\large \emph{Supervisor:} Prof. Eleni Vasilaki}\\[1cm] % if applicable
\large A report submitted in partial fulfilment of the requirements\\ for the degree of MSc
Data Analytics in Computer Science\\[0.3cm] 
\textit{in the}\\[0.3cm]
Department of Computer Science\\[2cm]
\today
\end{center}

\end{titlepage}

% -------------------------------------------------------------------
% Declaration
% -------------------------------------------------------------------

\newpage
\section*{\Large Declaration}

All sentences or passages quoted in this document from other people's work have been specifically acknowledged by clear cross-referencing to author, work and page(s).  Any illustrations that are not the work of the author of this report have been used with the explicit permission of the originator and are specifically acknowledged.  I understand that failure to do this amounts to plagiarism and will be considered grounds for failure.\\[1cm]

\noindent Name: Yue Zheng \\[1mm]
\rule[1em]{25em}{0.5pt}

% \noindent Signature:\\[1mm]
% \rule[1em]{25em}{0.5pt}

% \noindent Date:\\[1mm]
% \rule[1em]{25em}{0.5pt}

% -------------------------------------------------------------------
% Abstract
% -------------------------------------------------------------------

\chapter*{\Large \center Abstract}

% Guidance of how to write an abstract/summary provided by Nature: https://cbs.umn.edu/sites/cbs.umn.edu/files/public/downloads/Annotated_Nature_abstract.pdf

Reinforcement learning has exceeded human-level performance in game playing AI with deep learning methods according to the experiments from DeepMind on Go and Atari games. Deep learning solves high dimension input problems which stop the development of reinforcement for many years. This study uses both two techniques to create several agents with different algorithms that successfully learn to play T-rex Runner. Deep Q network algorithm and three types of improvements are implemented to train the agent. The results from some of them are far from satisfactory but others are better than human experts. Batch normalization is a method to solve internal covariate shift problems in deep neural network. The positive influence of this on reinforcement learning has also been proved in this study.

% -------------------------------------------------------------------
% Acknowledgement
% -------------------------------------------------------------------

\chapter*{\Large \center Acknowledgement}

I would like to express many thanks to my supervisor, Prof. Dr. Eleni
Vasilaki for assigning me this project and her guidance across this study. I would also like to acknowledge my dear friends for helping me to solve different problems and giving me inspiration. This include Mwiza L Kunda, Wei Wei, Yuliang Li, Ziling Li, Zixuan Zhang. Finally, I would like to acknowledge my family as well as my girl friend Fangni Liu for their encouragement during my project.

% -------------------------------------------------------------------
% Contents, list of figures, list of tables
% -------------------------------------------------------------------

\tableofcontents
\listoffigures
\listoftables

% -------------------------------------------------------------------
% Main sections (as required)
% -------------------------------------------------------------------

\mainmatter

\chapter{Introduction}

\section{Background}
The applications of Artificial Intelligence are widely used in recent years. As one part of them, Reinforcement Learning has achieved incredible results in game playing. An intelligent agent will be created and trained with reinforcement learning algorithms to fulfill this tasks. In the Future of Go Summit 2017, Alpha Go which is an AI player trained with deep reinforcement learning algorithms won three games against the world best human player in Go. The success of reinforcement learning in this area shock the world and many researches are launched such as driverless cars. Deep learning methods such as convolutional neural network contributes a lot to this because these techniques solves the problem of dealing with high dimension input data and feature extraction.

T-rex Runner is a dinosaur game from Google Chrome offline mode. The aim of the player is to escape all obstacles and get higher score until reaching the limitation which is $99999$. The moving speed of the obstacles will increase as time goes by which make it difficult to get the highest score. 

The code of this project can be found in \href{https://github.com/yzheng51/rl-dino-run}{this link} which is written in Python.

\section{Aim of the project}

The aim of this project is to create an agent using different algorithms to play T-rex Runner and compare the performance of them. Internal covariate shift is the change of distribution in each layer of during the training which may result in longer training time especially in deep neural network. To cope with this problem, batch normalization use linear transformation on each feature to normalize the data with the same mean and variance. The same problem may also occur in deep reinforcement learning because the decision is based on neural network. Beyond the comparison of different reinforcement learning algorithms, this project will also investigate the effect of batch normalization. The overall objectives of this project are list below.

\begin{itemize}
\item Create an agent to play T-rex Runner
\item Compare the difference among different reinforcement learning algorithms
\item Investigate the effect of batch normalization in reinforcement learning
\end{itemize}

\section{Overview}

This study opens with a literature review on deep learning and reinforcement learning. Each section includes the history of the field and the techniques related to this study. Chapter 3 includes the description of the game and the choice of algorithms according the literature review. The entire processing step will be shown as well as the architecture of the model. The design of the experiments and the evaluation methods are presented in this chapter too. Chapter 4 shows the result of all the experiments and the discussion of each experiment. Chapter 5 presents the conclusion of this study and the proposed future works.
\chapter{Literature Survey}

This chapter introduces the techniques used in developing an agent to play T-rex Runner. There are two main sections which are Deep Learning and Reinforcement Learning. Brief history and some milestones will be described and some important methods will be shown in detail.

\section{Deep Learning}

Deep learning is a class of Machine Learning model based on Artificial Neural Network (ANN). There two kinds of deep learning model which is widely used in recent years. Recurrent Neural Network is one of them which shows its power in Natural Language Processing. The other one plays an important role in deep reinforcement learning called Convolutional Neural Network (CNN). It is one of the most effective models for computer vision problems such as object detection and image classification. This section gives a brief introduction of deep learning and detailed information about convolutional neural network.

\subsection{History of Deep Learning}

An artificial neural network is a computation system inspired by biological neural networks which were first proposed by McCulloch, a neurophysiologist \cite{mcculloch1943logical}. In 1957, Perceptron was invented by Frank \cite{rosenblatt1957perceptron}. Three years later, his experiments show this algorithm can recognize some of alphabets \cite{rosenblatt1960perceptron}. However, Marvin proved that a single layer perceptron cannot deal with XOR problem \cite{minsky1969perceptron}. This stopped the development of ANN until Rumelhart et al. show that some useful representations can be learned with multi-layer perceptron, which is also called neural network, and backpropagation algorithm \cite{rumelhart1988learning} in 1988. One year later, LeCun et al. first used a five-layer neural network and backpropagation to solved digit classification problem and achieved great results \cite{lecun1998gradient}. His innovative model is known as LeNet which is the beginning of the convolutional neural network.

The origin of CNN was proposed by Fukushima named Neocognitron which was a self-organized neural network model with multiple layers \cite{fukushima1980neocognitron}. This model achieved a good result in object detection tasks because it is not position-sensitive. As mentioned before, LeCun et al. invented LeNet and got less than $1\%$ error rate in mnist handwritten digits dataset in 1998\cite{lecun1998gradient}. The model used convolutions and sub-sampling which is called convolution layer and pooling layer today to convert the original images into feature vectors and perform classification with fully connected layers. At the same time, some neural network models show some acceptable results in face recognition \cite{lawrence1997face}, speech recognition \cite{waibel1995phoneme} and object detection \cite{vaillant1994original}. But the lack of reliable theory caused the research of CNN to stagnate for many years.

In the ImageNet Large Scale Visual Recognition Challenge (ILSVRC) 2012 \cite{deng2009imagenet}, Alex and his team got $16.4\%$ error rate with an eight-layer deep neural network (AlexNet) \cite{krizhevsky2012imagenet}. This was a significant result compared with the one from second rank participant which was $26.2\%$. Beyond LeNet, AlexNet used eight layers to train the classifier with data augmentation, dropout, ReLU, which mitigated overfitting problem. Another significant discovery was that parallel computing with multiple GPUs can largely decrease the training time.

Two years later, Simonyan and Zisserman introduced a sixteen-layer neural network (VGGNet) and won the first prize in ILSVRC 2014 classification and localization tasks \cite{simonyan2014very}. This model got the state-of-the-art result with $7.3\%$ error rate at that time. VGGNet also proved that using smaller filter size and deeper network can improve the performance of CNN. The size of all filters in the model was no greater than $3\times 3$ while the first two layers in AlexNet were $11\times 11$ and $5\times 5$. In the same year, GoogLeNet \cite{szegedy2015going}, the best model of ILSVRC 2014 classification and detection tasks, first used inception which was proposed by Lin \cite{lin2013network} to solve vanishing gradient problem. Inception replaced one node with a network which was consisted of several convolutional layers and pooling layers then concatenate them before passing to the next layer. This change made the feature selection between two layers more flexible. In other words, it can be updated by the backpropagation algorithm.

Another problem of the deep neural network was degradation resulting in high training error caused by optimization difficulty. To solve that problem, He et al. proposed a deep residual learning framework (ResNet) using a residual mapping instead of stacking layers directly \cite{he2015resnet}. This model won the championship in ILSVRC 2015 with only $3.57\%$ error rate. His experiments show that this new framework can not only solve degradation problems but also can improve the computing efficiency. ResNet There were many variants based on ResNet such as Inception-ResNet \cite{szegedy2017inception} and DenseNet \cite{huang2017densely}. The former one combined improved inception techniques into ResNet. Every two convolutional layers were connected in the later model. This change mitigated vanishing gradient problems and improved the propagation of features.

\subsection{Deep Neural Network and Activation Function}\label{sec:dnn}

A neural network or multi-layer perceptron consists of three main components: the input layer, the hidden layer, and the output layer. Each unit in one layer called a neuron. The input data are fed into the input layer conducting linear transformation through weights in the hidden layer. Finally, the result will be given non-linear ability through activation function and fed into the output layer. 

\begin{figure}[ht]
\centering
\includegraphics[width=15cm]{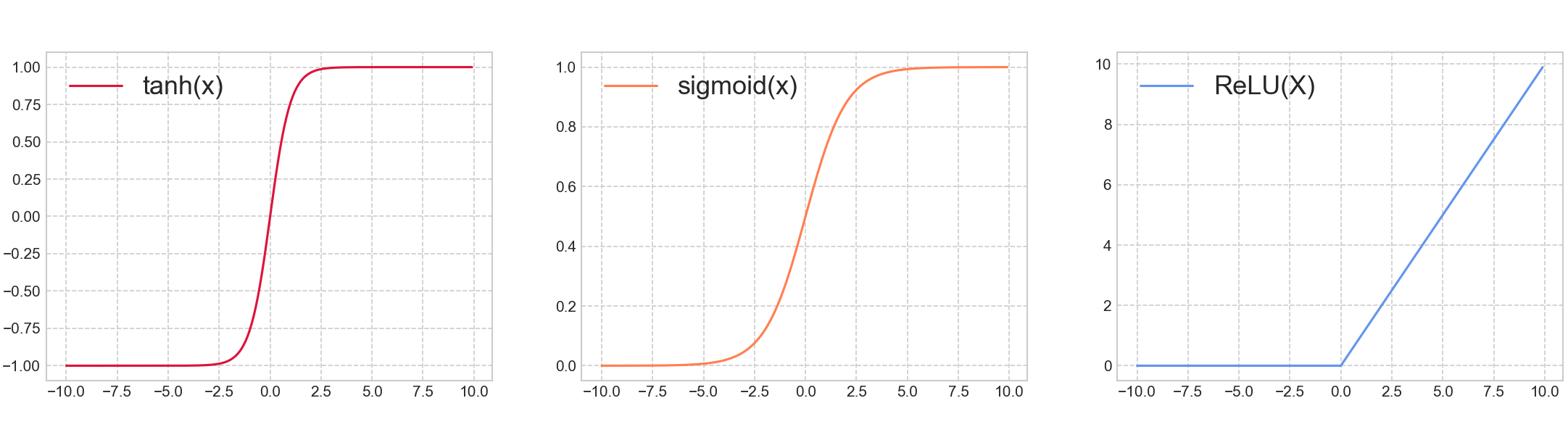}
\caption{Three type of activation functions.}
\label{fig:activation}
\end{figure}

Activation function enables the network to learn more complicated relationships between inputs and outputs. There are three widely used activation functions shown in Figure \ref{fig:activation}: sigmoid, tanh and ReLU. ReLU is the most commonly used one in three because it has a low computational requirement and better performance in solving vanishing gradient problems compared with the other two.

\begin{equation}
\text{tanh}(x) = \frac{e^x-e^{-x}}{e^x+e^{-x}}
\label{equ:tanh}
\end{equation}

\begin{equation}
\text{sigmoid}(x) = \frac{1}{1+e^{-x}}
\label{equ:sigmoid}
\end{equation}

\begin{equation}
\text{ReLU}(x) = \max(x, 0)
\label{equ:relu}
\end{equation}

\begin{figure}[ht]
\centering
\includegraphics[width=15cm]{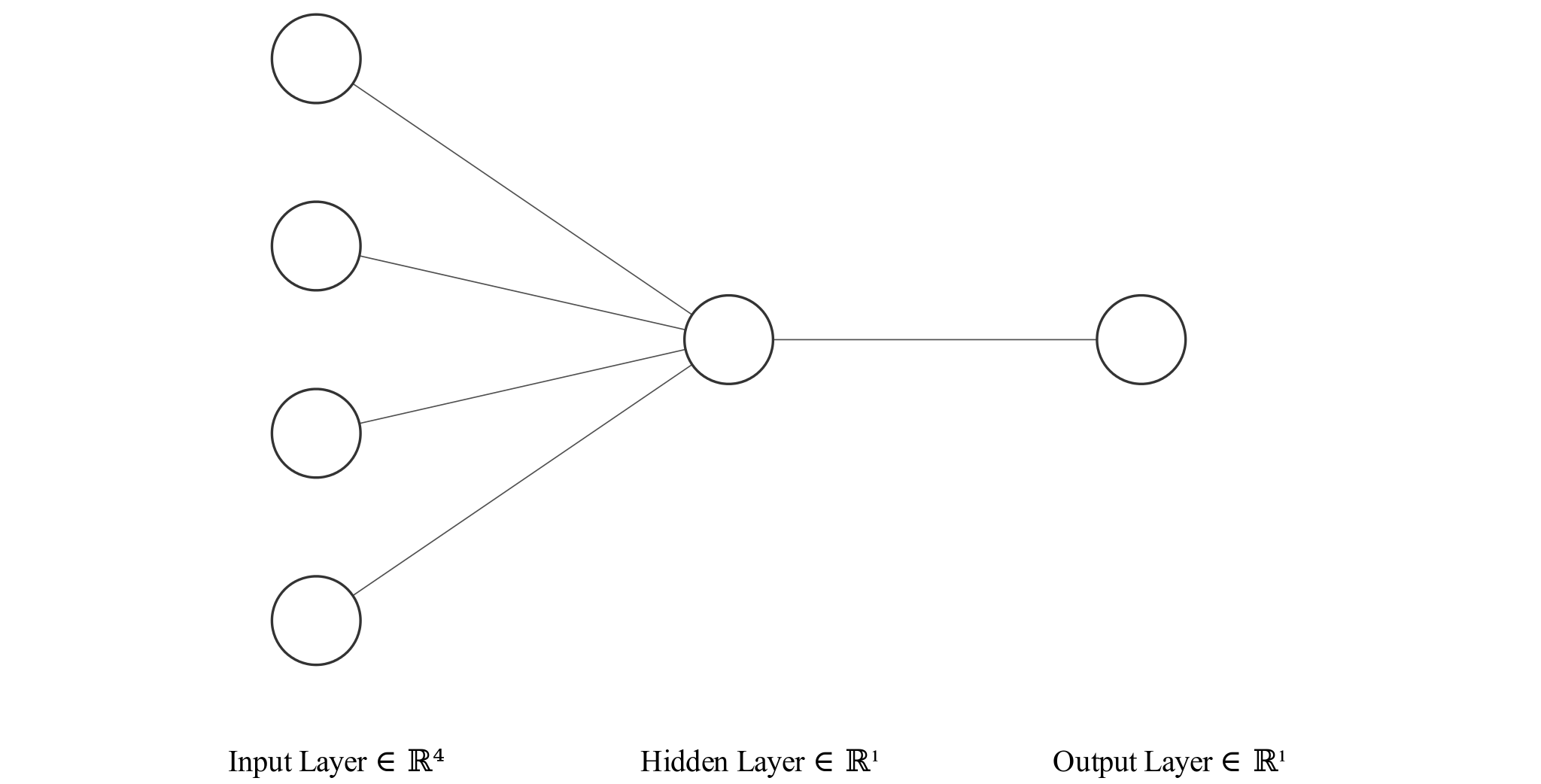}
\caption{A simple neural network.}
\label{fig:nn}
\end{figure}

To illustrate the entire process in neural network, here is an example in Figure \ref{fig:nn}. Given an input data $T=\{(\boldsymbol{x}_1, y_1),(\boldsymbol{x}_2,y_2),(\boldsymbol{x}_3,y_3),(\boldsymbol{x}_4,y_4)\,|\,\boldsymbol{x}_i\in\mathbb{R}^m\}$ and a randomly generated weight $[\boldsymbol{w}_1,\boldsymbol{w}_2,\boldsymbol{w}_3,\boldsymbol{w}_4]^T$ in the hidden layer, the output of the neural network is $\hat{y}$ and the activation of the hidden layer is $f$. Therefore, the estimated value $\hat{y}_i$ can be calculated by

\begin{figure}[ht]
\centering
\includegraphics[width=15cm]{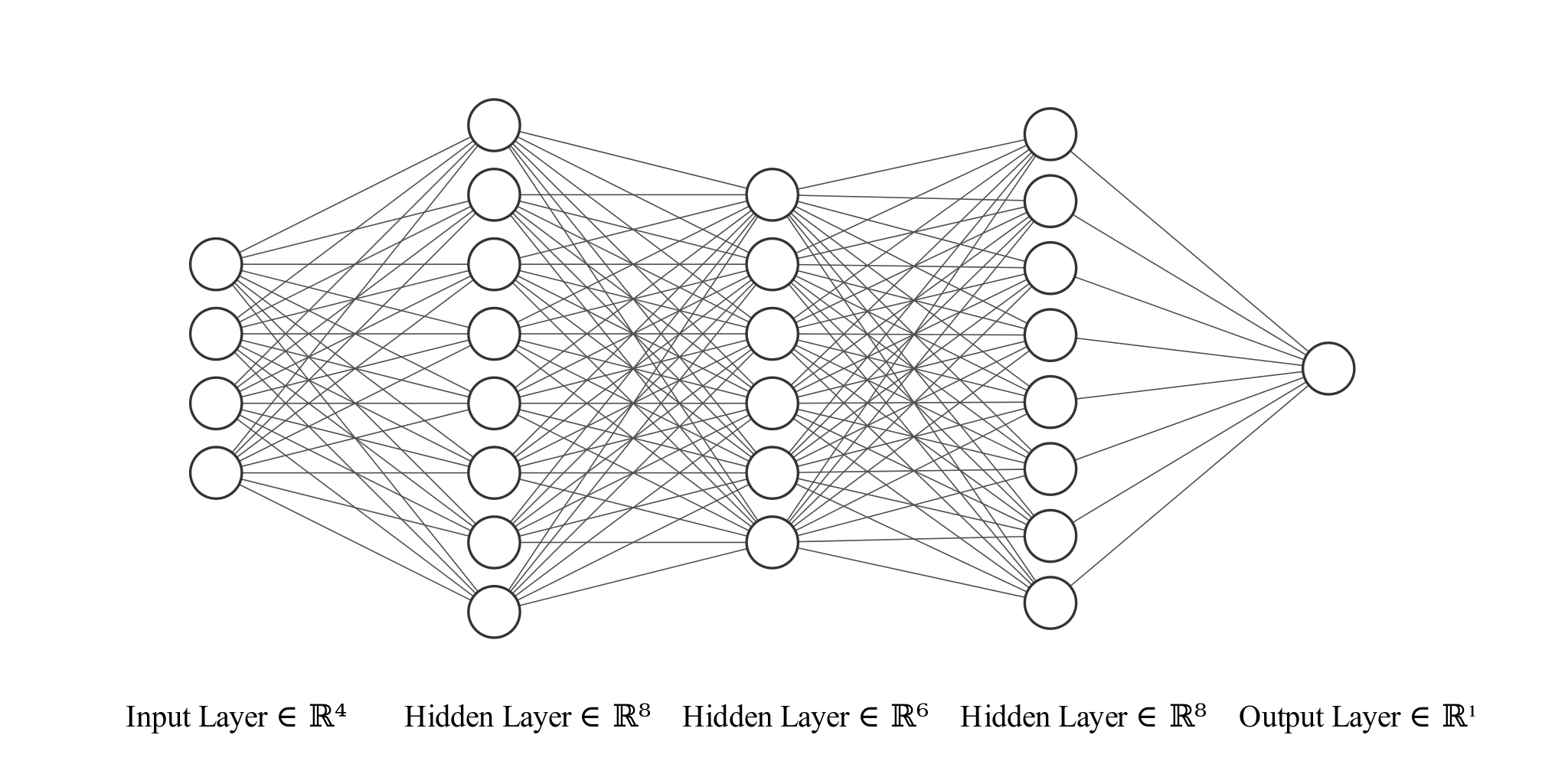}
\caption{A simple deep neural network with three hidden layers.}
\label{fig:dnn}
\end{figure}

\begin{equation}
\hat{y}_i=f(\boldsymbol{w}_i\boldsymbol{x}_i+b)
\end{equation}

where $b$ is the bias of the hidden layer and $m$ is the number of features. If the number of hidden layers in the neural network is greater than two, this is also called Deep Neural Network (DNN). Consider a simple DNN with three hidden layers shown in Figure \ref{fig:dnn}. Given the same input $\boldsymbol{X}=[\boldsymbol{x}_1,\boldsymbol{x}_2,\boldsymbol{x}_3,\boldsymbol{x}_4]^T$ in matrix form and $\boldsymbol{W}_i$ is the weight between $(i-1)$-th and $i$-th layer, the output of $i$-th layer $\boldsymbol{a}_i$ can be calculated by

\begin{equation}
\boldsymbol{a}_i=f_i(\boldsymbol{z}_i)
\label{equ:dnn}
\end{equation}

where $f_i$ is the activation function between $(i-1)$-th and $i$-th layer and $\boldsymbol{z}_i=\boldsymbol{a}_{i-1}\boldsymbol{W}_i$, especially $\boldsymbol{a}_0=\boldsymbol{X}$. The dimension of those variables are shown in Table \ref{tab:dnn-dimension}

\begin{table}[ht]
\centering
\begin{tabular}{llc}
\hline\hline
\textbf{Parameter} & \textbf{Description} & \textbf{Dimension}\\
\hline
$\boldsymbol{X}$ & Input data & $4\times m$ \\
$\boldsymbol{W}_1$& Weight between input layer and  hidden layer 1 & $m\times 8$\\
$\boldsymbol{W}_2$& Weight between hidden layer 1 and hidden layer 2 & $8\times 6$\\
$\boldsymbol{W}_3$& Weight between hidden layer 2 and hidden layer 3 & $6\times 8$ \\
$\boldsymbol{W}_4$& Weight between hidden layer 3 and output layer  & $8\times 1$ \\
\hline
\end{tabular}
\caption{Dimension description of the deep neural network}
\label{tab:dnn-dimension}
\end{table}

\subsection{Backpropagation Algorithm}\label{sec:backprop}

Section \ref{sec:dnn} introduces the way to estimated label using deep neural network. In order to optimize this estimation, a cost function $J$ is used to quantify the difference between the estimated value $\hat{y}$ and the true value $y$. To give a simple example, Mean Square Error \cite{allen1971mean} which is often applied to regression problems is used in this section to illustrate how the backpropagation algorithm works. Equation \ref{equ:mse} shows the form of mean square error.

\begin{equation}
J(w,x) = \frac{1}{2}\sum_{i=1}^n (y_i-\hat{y}_i)^2
\label{equ:mse}
\end{equation}

where $x$ is the input data and $w$ represent all weights used in the model. Thus, the optimization problem can be described as following

\begin{equation}
\min\limits_{w} J(w,x)
\label{equ:sgd-objective}
\end{equation}

Stochastic Gradient Descent (SGD) is an effective way to solve this optimization problem if $J(w,x)$ is convex. However, it still shows acceptable results in deep neural network even though there is no guarantee for global optimal point in non-convex optimization \cite{ge2015escaping}. Instead of finding the optimal point directly, SGD optimize the objective function \ref{equ:sgd-objective} iteratively by following equation

\begin{equation}
w\leftarrow w - \eta\frac{\partial J(w,x)}{\partial w}
\label{equ:sgd-formula}
\end{equation}

where $\eta$ is the learning rate which can control the update speed of the weight. Using gradient methods such as SGD to optimize the cost function in neural network is called backpropagation algorithm. Considering the deep neural network shown in \ref{fig:dnn} and the techniques of matrix calculus \cite{hu2012matrix}, the gradient of $J$ with respect to $\boldsymbol{W}_4$ is

\begin{equation}
\frac{\partial J}{\partial \boldsymbol{W}_4}=\boldsymbol{a}_3^T\left((\boldsymbol{a}_4-y)\odot\nabla f_4(\boldsymbol{z}_4)\right)
\end{equation}

where $\odot$ is element-wise matrix multiplication and $\nabla f_4(\boldsymbol{z}_4)$ is the gradient with respect to $\boldsymbol{z}_4$. Results for other $\boldsymbol{W}_i$ can be calculated in a similar way. With equation \ref{equ:sgd-formula}, $\boldsymbol{W}_i$ can be updated during each iteration by

\begin{equation}
\boldsymbol{W}_i\leftarrow \boldsymbol{W}_i - \eta\frac{\partial J}{\partial \boldsymbol{W}_i}
\end{equation}

\subsection{Convolutional Neural Network}

Compared with a common deep neural network, a convolutional neural network has two extra components which are convolutional layer and pooling layer. The convolutional layers make use of several trainable filters to select different features. The pooling layer reduces the dimension of the data by subsampling. 

\begin{figure}[ht]
\centering
\includegraphics[width=12cm]{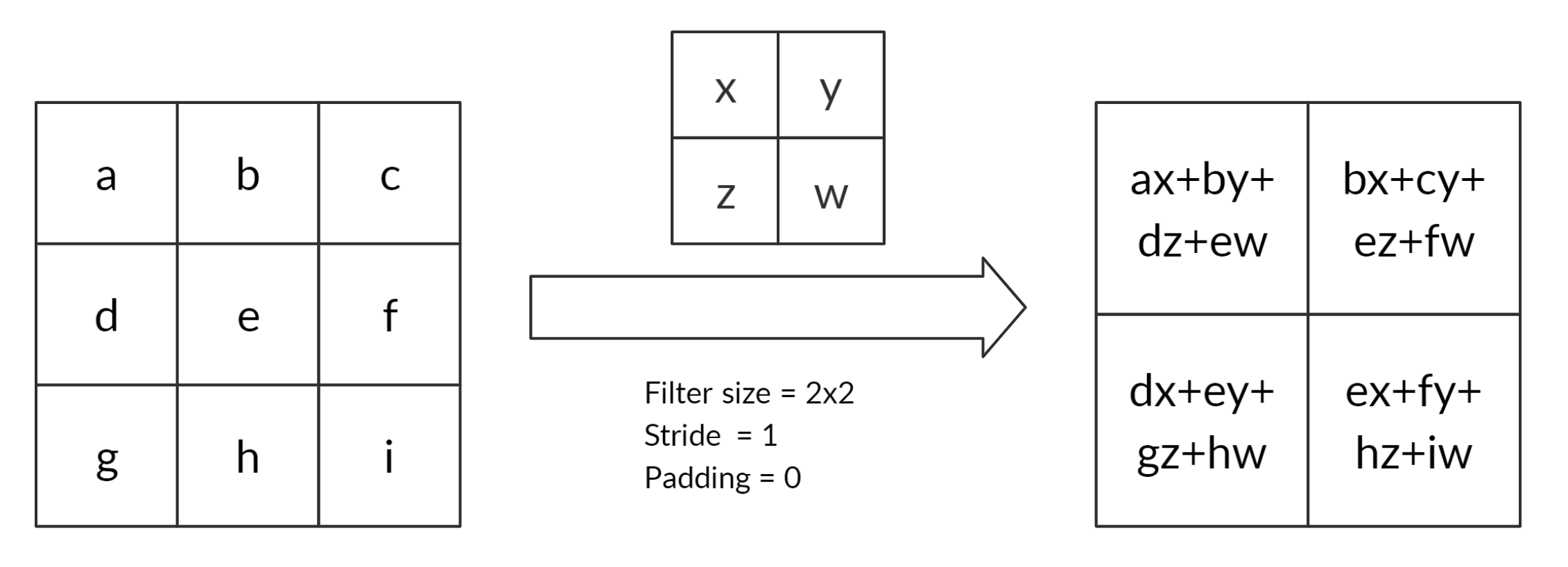}
\caption{Operations in convolution layer.}
\label{fig:con-layer}
\end{figure}

In the convolution layer, the output of the last layer is convolved by trainable filters with element-wise matrix multiplication. The size and the number of each filter are defined by the user and the initial value is randomly generated. The moving step of a filter in each convolution layer is decided by stride. In order to keep the information of the border during the forward propagation, a series of zeros attached to the border of the image called padding. Figure \ref{fig:con-layer} shows how the result of one neuron in a convolution layer comes from and how the filter $(x,y,z,w)$ will be updated in every iteration in backpropagation.

\begin{figure}[ht]
\centering
\includegraphics[width=12cm]{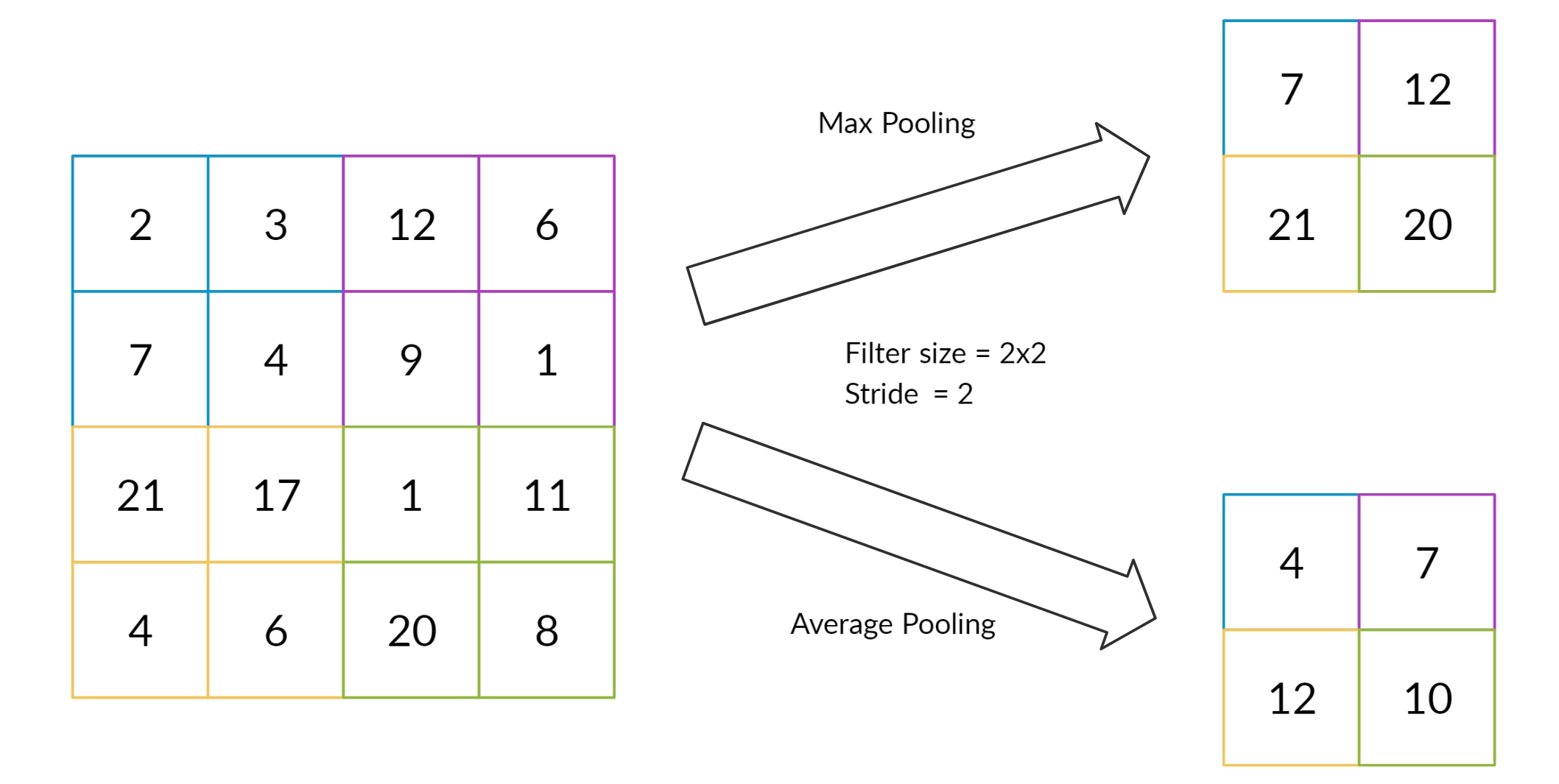}
\caption{Operations in pooling layer.}
\label{fig:pool-layer}
\end{figure}

The reason for using a pooling layer is not only for dimension reduction but also for detecting invariant features including translation, rotation, scale from the input \cite{scherer2010evaluation}. There are two types of operations in pooling layer: max pooling and average pooling. In max pooling, only the maximum value in user-defined windows will be chosen while all values in the window will make contributions to the output in average pooling. The choice of operation is dependent on tasks and Boureau has made a theoretical comparison between those two \cite{boureau2010learning}. Both max and average operation are shown in Figure \ref{fig:pool-layer}.

\begin{figure}[ht]
\includegraphics[width=15cm]{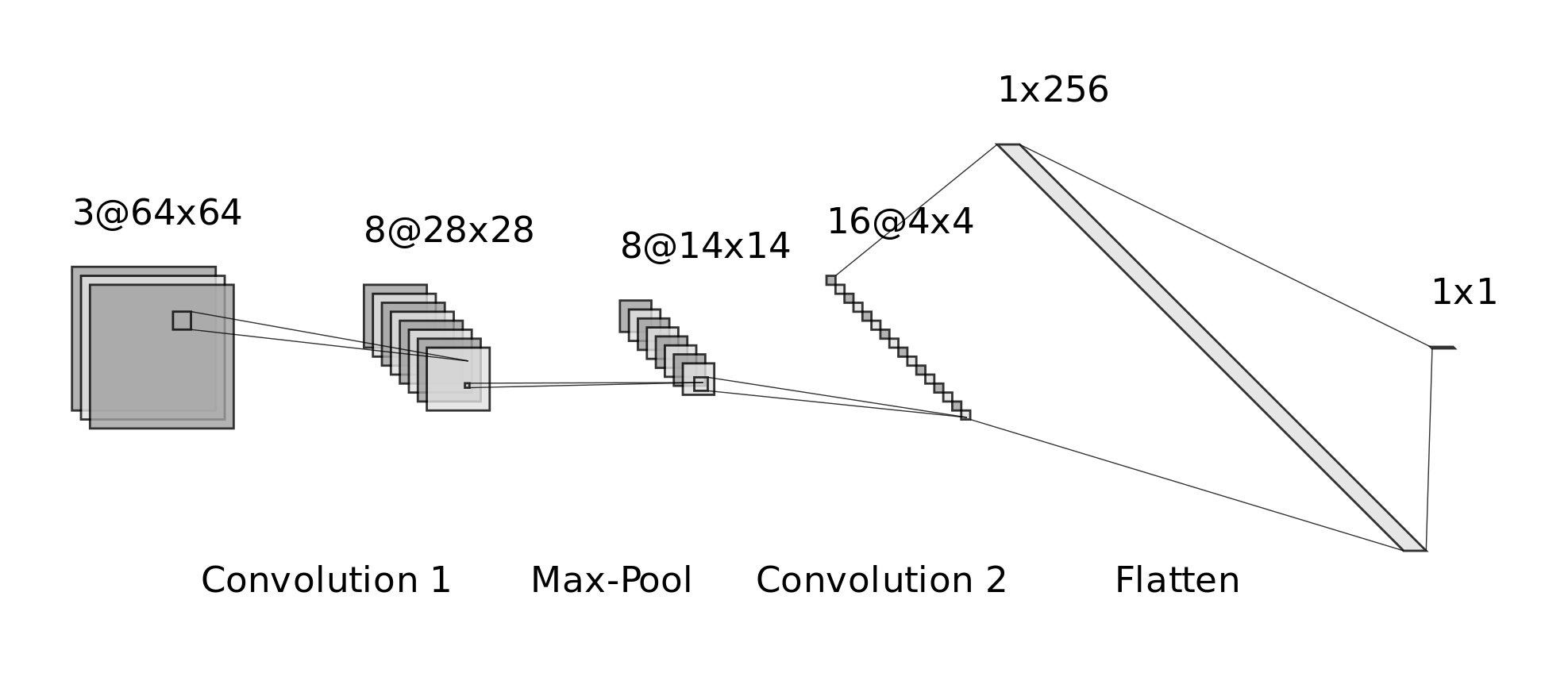}
\caption{A simple convolutional neural network.}
\label{fig:cnn}
\end{figure}

A complete convolutional neural network consists of several convolutional layers, pooling layers, and fully connected layers. The fully connected layer is the same concept of DNN which used the flatten vector of the last output of the other two layers as input. Considering a classification problem as shown in Figure \ref{fig:cnn}, an image with a size of $64\times 64$ is fed into CNN and output a scalar which represents its class. Table \ref{tab:cnn-info} lists the filters information used in CNN.

\begin{table}[ht]
\centering
\begin{tabular}{cccccc}
\hline\hline
\textbf{Layer} & \textbf{Numbers} & \textbf{Size} & \textbf{Stride} & \textbf{Padding} & \textbf{Output dimension} \\
\hline
Convolution 1 & 8 & 8$\times$8 & 2 & 0 & 8$\times$28$\times$28\\
Max Pooling & 8 & 2$\times$2 & 2 & / & 8$\times$14$\times$14\\
Convolution 2 & 16 & 6$\times$6 & 2 & 0  & 16$\times$4$\times$4\\
\hline
\end{tabular}
\caption{Property of convolutional layer and pooling layer}
\label{tab:cnn-info}
\end{table}

The backpropagation algorithm in convolutional neural network is a little different from described in section \ref{sec:backprop} because of two extra layer type. In average pooling layer, the error will be divided by $t\times t$ which is the size of the filter and propagate to the last layer. In max pooling layer, the position of the maximum value will be stored when forward propagating and the error will be directly passed through that position. In convolutional layer, backpropagation can be calculated through basic differentiation. Consider the convolution operation in Figure \ref{fig:con-layer}, if the error from the output layer is $\delta$, then we have
\begin{equation}
\frac{\partial \delta}{\partial x}=\frac{\partial \delta}{\partial O_{11}} \frac{\partial O_{11}}{\partial x}+\frac{\partial \delta}{\partial O_{12}} \frac{\partial O_{12}}{\partial x}+\frac{\partial \delta}{\partial O_{21}} \frac{\partial O_{21}}{\partial x}+\frac{\partial \delta}{\partial O_{22}} \frac{\partial O_{22}}{\partial x}
\end{equation}

where

\begin{equation}
\left[\begin{array}{ll}{O_{11}} & {O_{12}} \\ {O_{21}} & {O_{22}}\end{array}\right]
\end{equation}

is the output matrix in Figure \ref{fig:con-layer}. The differentiation of $\delta$ with respect to $y$, $z$, $w$ can be computed in a similar way.

\subsection{Batch Normalization} \label{sec:bn}

With the increasing depth of the neural network, the training time becomes longer. One of the reason is the distribution of input in each layer changes when updating the weight which is called Internal Covariate Shift. In 2015, Ioffe proposed Batch Normalization (BN) which make the distribution in each layer more stable and achieve shorter training time \cite{ioffe2015batch}. In each neuron, the input can be normalized by Equation \ref{equ:normalization} 

\begin{equation}
\widehat{x}=\frac{x-\mathbb{E}[x]}{\sqrt{\operatorname{Var}[x]+\epsilon}}
\label{equ:normalization}
\end{equation}

where $\epsilon$ is used to avoid zero variance. Now data in each neuron follow the distribution with mean $0$ and standard deviation $1$. However, this changes the representation ability of the network which may lead to the loss of information in the earlier layer. Therefore, Ioffe used another linear transformation to restore that representation

\begin{equation}
\tilde{x}=m \widehat{x} + n
\end{equation}

where $m$ and $n$ are learnable parameters, especially, the result is the same as original when $m=\sqrt{\operatorname{Var}[x]}$ and $n=\mathbb{E}[x]$. The mean and variance during training will be stored and will be treated as the mean of the variance of test data. In their experiment, BN can not only deal with Internal Covariate Shift problems but also mitigate vanishing gradient problems.

\section{Reinforcement Learning}

\begin{figure}[ht]
\centering
\includegraphics[width=15cm]{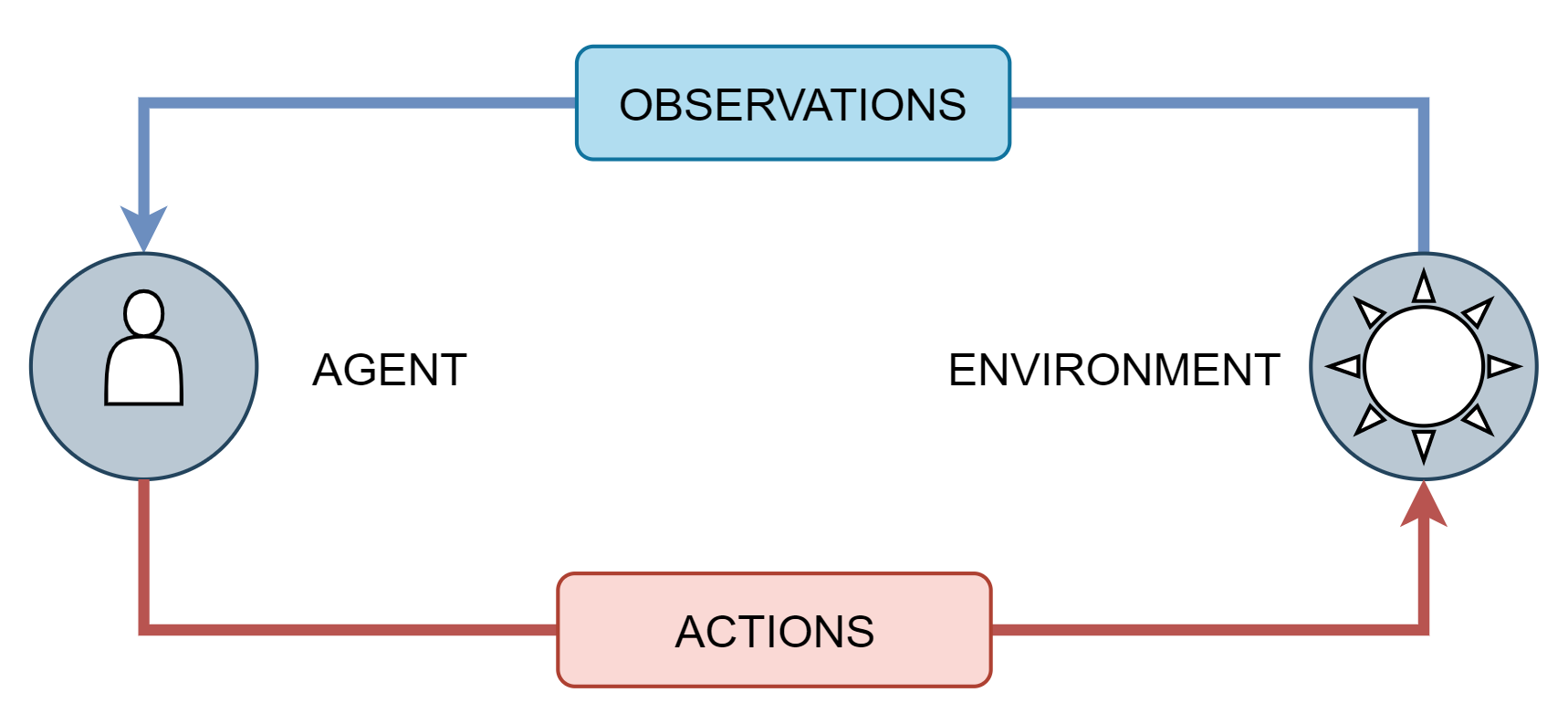}
\caption{Interaction between the agent and the environment.}
\label{fig:rl-env}
\end{figure}

Reinforcement Learning (RL) is a class of machine learning aiming at maximum the reward signal when making decisions. The basic component of reinforcement learning is the agent and the environment. As shown in Figure \ref{fig:rl-env}, the agent will receive feedback including observation and reward from the environment after each action. To generate a better policy, it will keep interacting with the environment and improve its decision-making ability step by step until the policy converges.

\subsection{History of Reinforcement Learning}\label{sec:rl}

In recent years, reinforcement learning becomes popular because of Alpha Go, a program that can beat human expert in Go \cite{silver2017mastering}. In the Future of Go Summit 2017, Alpha Go Master shocked the world by winning all three games against Ke Jie, the world best player in Go. But the research of reinforcement learning started very early. According to Sutton, the early history of RL can be divided into two main threads \cite{sutton2018reinforcement}.

One of them was optimal control. To cope with arising optimal control problems which were called "multi-stage decision processed" in 1954, the theory Dynamic Programming (DP) was introduced by Bellman \cite{bellman1954theory}. In the theory, he proposed the concept of "functional equation", which was often called the Bellman equation today. Although DP was one of the most effective approaches to solve optimal control problems at that time, the high computational requirements which is called "the curse of dimensionality" by Bellman were not easy to solve \cite{bellman1958combinatorial}. Three years later, he built a model called Markov Decision Processes (MDPs) to describe a kind of discrete deterministic processes \cite{bellman1957markov}. This deterministic system and the concept of value function which is described in the Bellman equation consists of the basic theory of modern reinforcement learning.

In optimal control thread, solving problems required full knowledge of the environment and it was not a feasible way to deal with most problems in the real world. The trial-and-error thread focused more on the feedback rather than the environment itself. The first expression about the key idea of trail-and-error including "selectional" and "associative" called "Law of Effect" was written in Edward Thorndike's book "Animal Intelligence" \cite{thorndike1911animal}. Although supervised learning was not "selectional", some researchers still mistook it for reinforcement learning and concentrated on pattern recognition \cite{clark1955generalization, widrow1960adaptive}. This led to rare researches in actual trial-and-error learning until Klopf recognized the difference between supervised learning and RL: the motivation to gain more rewards from the environment \cite{klopf1972brain, klopf1982hedonistic}. However, there were still some remarkable works such as the reinforcement learning rule called "selective bootstrap adaptation" by Widrwo in 1973 \cite{widrow1973punish}.

Both of two threads came across in modern reinforcement learning. Temporal Difference (TD) learning was a method that predicts future values depend on the current signal which originated from animal learning psychology. This idea was first proposed and implemented by Samuel \cite{samuel1959aerosol}. In 1972, Klopf developed the idea of "generalized reinforcement" and linked the trial-and-error learning with animal learning psychology \cite{klopf1972brain}. In 1983, Sutton developed and implemented the actor-critic architecture in trial-and-error learning based on the idea from Klopf \cite{barto1983neuronlike}. Five years later, he proposed TD($\lambda$) algorithms which used additional step information for update policy and made TD learning a general prediction method for deterministic problems \cite{sutton1988learning}. One year later, Chris used optimal control methods to solve temporal-difference problems and developed the Q-learning algorithm which estimated delayed reward by action value function \cite{watkins1989learning}. In 1994, an online Q-learning was proposed by Rummery and Niranjan which was known as SARSA \cite{rummery1994line}. The difference between Q-learning and SARSA was that the agent used the same policy during the learning process in SARSA while it always chooses the best action based on value function in Q-learning.

With the development of the deep neural network, DeepMind proposed Deep Q-learning Network (DQN) algorithm which used a convolutional neural network to solve high dimensionality of the state in reinforcement learning problems \cite{mnih2013playing}. Two years later, they modified DQN by adding a target policy to improve its stability \cite{mnih2015human}. The highlight of the DQN was not only the combination of deep learning and RL but also the experience replay mechanism. To solve dependency problems when optimizing CNN, Mnih et al. stored the experiences to memory in each step and randomly sampled a mini-batch to optimize the neural network based on the idea from Lin \cite{lin1992self}. In 2015, this mechanism was improved by measuring the importance of experience with temporal difference error \cite{schaul2015prioritized}. Meanwhile, Wang proposed Dueling DQN which used an advantage function learning how valuable a state was without estimating each action value for each state \cite{wang2015dueling}. This new neural network architecture was helpful when there was no strong relationship between actions and the environment. In 2016, DeepMind proposed Double DQN which show the higher stability of the policy by reducing overestimated action values \cite{van2016deep}.

Although a series of algorithms based on DQN show human-level performance on Atari games, they still failed to deal with some specific games. DQN was a value-based method which meant the choice of action was depend on the action values. However, choosing action randomly may be the best policy in some games such as Rock$-$paper$-$scissors. To deal with this problem, Sutton proposed policy gradient which enabled the agent to optimize the policy directly \cite{sutton2000policy}. Based on this, OpenAI proposed a new family of algorithms such as Proximal Policy Optimization (PPO) \cite{schulman2017proximal}. PPO used a statistical method called importance sampling which was used to estimate a distribution by sampling data from another distribution and this simple modification show a better performance in RoboschoolHumanoidFlagrun.

Since the basic policy gradient method sampled data from completed episodes, the variance of the estimation was high because of the high dimension action space. Similar to value-based method, Actor-critic method was proposed to solve this problem \cite{konda2000actor}. Compared with the policy gradient, this method used a critic to evaluate the chosen action. This made the policy can be updated after each decision which not only reduced the variance but also accelerated the convergence. The famous improved actor-critic based algorithm is asynchronous advantage actor-critic (A3C) \cite{mnih2016asynchronous}. Similar to Dueling DQN, this method used advantage function to estimate value function and performed computing in parallel which can largely increase the learning speed.

\subsection{Markov Decision Processes} \label{sec:markov-decision-process}

As mentioned in Section \ref{sec:rl}, the interaction between the agent and the environment can be modeled as a Markov Decision Process which is based on Markov property. Markov property describes a kind of stochastic processes that the probability of next event occurring only depend on the current event.

\begin{definition}[Markov property \cite{silver2015notes}]
Given a state $S_{t+1}$ at time $t+1$ in a finite sequence $\{S_0, S_1, S_2, \cdots, S_N\}$. This sequence has Markov property, if and only if
\begin{equation}
\mathbb{P}\left[S_{t+1} | S_{t}\right]=\mathbb{P}\left[S_{t+1} | S_{1}, \ldots, S_{t}\right]
\end{equation}
\label{def:markov-property}
\end{definition}

A Markov Decision Process (MDP) is a random process with Markove property, values and decisions.

\begin{definition}[Markov Decision Process \cite{silver2015notes}]
A Markov Decision Process can be described as a tuple $\langle\mathcal{S}, \mathcal{A}, \mathcal{P}, \mathcal{R}, \gamma\rangle$
\begin{itemize}
\item $\mathcal{S}$ is a finite set of states
\item $\mathcal{A}$ is a finite set of actions
\item $\mathcal{P}$ is a state transition probability matrix
\begin{equation}
\mathcal{P}_{s s^{\prime}}^{a}=\mathbb{P}\left[S_{t+1}=s^{\prime} | S_{t}=s, A_{t}=a\right]  
\end{equation}
\item $\mathcal{R}$ is a reward function
\begin{equation}
\mathcal{R}_{s}^{a}=\mathbb{E}\left[R_{t+1} | S_{t}=s, A_{t}=a\right]    
\end{equation}
\item $\gamma$ is a discount factor $\gamma \in[0,1]$
\end{itemize}
\label{def:markov-decision-process}
\end{definition}

To describe how the decision is made, a policy $\pi$ is required to define the behaviour of the agent.

\begin{definition}[Policy \cite{silver2015notes}]
A policy is a distribution over actions given states
\begin{equation}
\pi(a | s)=\mathbb{P}\left[A_{t}=a | S_{t}=s\right]
\end{equation}
\label{def:policy}
\end{definition}

In MDP, the agent is expected to get as many rewards as it can from the environment. However, maximizing the reward at time-step $t$ makes the agent short-sighted which means it only considers the reward from the next action rather the total reward of one episode. Therefore, return is defined as the concept "reward" which the agent is expected to maximize.

\begin{definition}[Return \cite{silver2015notes}]
The return $G_t$ is the total discounted reward $R_t$ from time-step $t$.
\begin{equation}
G_{t}=R_{t+1}+\gamma R_{t+2}+\ldots=\sum_{k=0}^{\infty} \gamma^{k} R_{t+k+1}
\label{equ:return}
\end{equation}
where $\gamma\in[0,1]$ is the discount factor.
\label{def:return}
\end{definition}

\begin{figure}[ht]
\centering
\includegraphics[width=15cm]{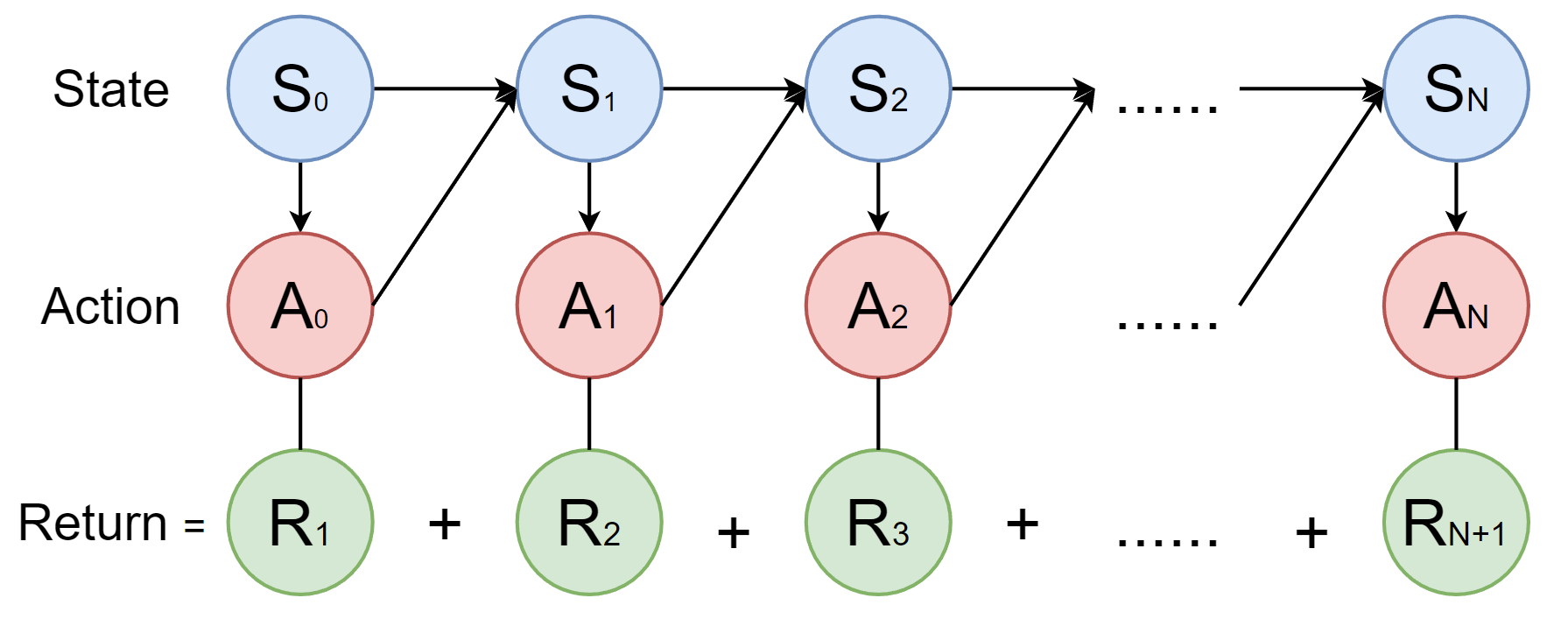}
\caption{Markov decision process in reinforcement learning.}
\label{fig:markov-decision-process}
\end{figure}

The value of the discount factor represents how far-sighted the agent will be. If this value is $1$, the agent will treat every reward in the future as the same. But this will also make the agent confused about which decision is not appropriate. At this point, the behavior of the agent can be described as in Figure \ref{fig:markov-decision-process}

Since the return is defined in a random process, similar to reward function, the expectation of it can be defined as following which is also called value function.

\begin{definition}[State Value Function \cite{silver2015notes}]
The state-value function $v_{\pi}(s)$ of an MDP is the expected return
starting from state $s$, and then following policy $\pi$
\begin{equation}
v_{\pi}(s)=\mathbb{E}_{\pi}\left[G_{t} | S_{t}=s\right]
\end{equation}
\label{def:state-value-function}
\end{definition}

\begin{definition}[Action Value Function \cite{silver2015notes}]
The action-value function $q_{\pi}(s, a)$ is the expected return
starting from state $s$, taking action a, and then following policy $\pi$
\begin{equation}
q_{\pi}(s, a)=\mathbb{E}_{\pi}\left[G_{t} | S_{t}=s, A_{t}=a\right]
\end{equation}
\label{def:action-value-function}
\end{definition}

With the Definition \ref{def:state-value-function}, \ref{def:action-value-function} and the definition of the expectation, we can simply write
\begin{equation}
v_{\pi}(s)=\sum_{a \in \mathcal{A}} \pi(a | s) q_{\pi}(s, a)
\label{equ:state-action-relation}
\end{equation}
where $\mathcal{A}$ is a set of action the agent can choose.

\subsection{Bellman Equation}

Since we define the Markov decision process in Section \ref{sec:markov-decision-process}, the behavior of the agent can be described mathematically. As mentioned in Section \ref{sec:rl}, this problem can be solved by the Bellman equation.

\begin{theorem}[Bellman Expectation Equation \cite{silver2015notes}]
The state-value function can be decomposed into immediate reward plus discounted value of successor state
\begin{equation}
v_{\pi}(s)=\mathbb{E}_{\pi}\left[R_{t+1}+\gamma v_{\pi}\left(S_{t+1}\right) | S_{t}=s\right]
\label{equ:bellman-expectation-state}
\end{equation}

The action-value function can similarly be decomposed
\begin{equation}
q_{\pi}(s, a)=\mathbb{E}_{\pi}\left[R_{t+1}+\gamma q_{\pi}\left(S_{t+1}, A_{t+1}\right) | S_{t}=s, A_{t}=a\right]
\label{equ:bellman-expectation-action}
\end{equation}
\label{def:bellman-expectation-equation}
\end{theorem}

Here is a simple proof for Equation \ref{equ:bellman-expectation-action}. According to the Definition \ref{def:return}, the return at time $t$ can be decomposed into two parts: the immediate reward and the discounted return at time $t+1$
\begin{equation}
G_t=R_{t+1}+\gamma G_{t+1}
\label{equ:discounted-return}
\end{equation}

Substitute $G_t$ with Equation \ref{equ:discounted-return} in Definition \ref{def:action-value-function}

\begin{equation}
q_{\pi}(s, a)=\mathbb{E}_{\pi}\left[R_{t+1}+\gamma G_{t+1} | S_{t}=s, A_{t}=a\right]
\label{equ:proof-bellman-expectation-action}
\end{equation}

Due to the linearity of expectation, $G_{t+1}$ can be replaced by $q_{\pi}\left(S_{t+1}, A_{t+1}\right)$ and then we obtain the Bellman equation for action-value function. The state-value function can be proved in the same way. With the definition of optimal value function

\begin{definition}[Optimal Value Function \cite{silver2015notes}]
The optimal state-value function $v_{*}(s)$ is the maximum value function over all policies
\begin{equation}
v_{*}(s)=\max _{\pi} v_{\pi}(s)
\end{equation}
The optimal action-value function $q_{*}(s, a)$ is the maximum action-value function over all policies
\begin{equation}
q_{*}(s, a)=\max _{\pi} q_{\pi}(s, a)
\end{equation}
\end{definition}

Theorem \ref{def:bellman-expectation-equation} can be extended to Bellman optimality equation

\begin{theorem}[Bellman Optimality Equation \cite{silver2015notes}]
The optimal state-value function can be decomposed into maximum immediate reward plus discounted optimal value of successor state
\begin{equation}
v_{*}(s)=\max _{a} \mathcal{R}_{s}^{a}+\gamma \sum_{s^{\prime} \in \mathcal{S}} \mathcal{P}_{s s^{\prime}}^{a} v_{*}\left(s^{\prime}\right)
\label{equ:bellman-optimality-value}
\end{equation}
The optimal action-value function can similarly be decomposed
\begin{equation}
q_{*}(s, a)=\mathcal{R}_{s}^{a}+\gamma \sum_{s^{\prime} \in \mathcal{S}} \mathcal{P}_{s s^{\prime}}^{a} \max _{a^{\prime}} q_{*}\left(s^{\prime}, a^{\prime}\right)
\label{equ:bellman-optimality-action}
\end{equation}
where $s=S_t$, $a=A_t$, $s^\prime=S_{t+1}$, $a^\prime=A_{t+1}$
\end{theorem}

Here is a simple proof for Equation \ref{equ:bellman-optimality-action}. Due to the linearity of expectation, Equation \ref{equ:proof-bellman-expectation-action} can be decomposed into the expectation of the immediate reward
\begin{equation}
\mathbb{E}_{\pi}\left[R_{t+1} | S_{t}=s, A_{t}=a\right]
\label{equ:proof-bellman-reward}
\end{equation}

and the expectation of the discounted return at time $t+1$

\begin{equation}
\gamma\mathbb{E}_{\pi}\left[G_{t+1} | S_{t}=s, A_{t}=a\right]
\label{equ:proof-bellman-discounted-return-1}
\end{equation}

According to the definition of reward function in Definition \ref{def:markov-decision-process}, Equation \ref{equ:proof-bellman-reward} is equal to $\mathcal{R}_{s}^{a}$. If next state is $s^{\prime}$, Equation \ref{equ:proof-bellman-discounted-return-1} can be written as following with the transition probability matrix $\mathcal{P}_{s s^{\prime}}^{a}$

\begin{equation}
\gamma\sum_{s^{\prime} \in \mathcal{S}} \mathcal{P}_{s s^{\prime}}^{a}\mathbb{E}_{\pi}\left[G_{t+1} | S_{t}=s, A_{t}=a, S_{t+1}=s^{\prime}\right]
\label{equ:proof-bellman-discounted-return-2}
\end{equation}

With the Markov property, we know the expectation of the return in Equation \ref{equ:proof-bellman-discounted-return-2} is not related to the current state $s$ and action $a$ and this is equal to the state-value function. Therefore, Equation \ref{equ:proof-bellman-discounted-return-2} can be written as following
\begin{equation}
\gamma\sum_{s^{\prime} \in \mathcal{S}} \mathcal{P}_{s s^{\prime}}^{a}v(s^{\prime})
\label{equ:proof-bellman-discounted-return-3}
\end{equation}

Considering \ref{equ:proof-bellman-reward}, \ref{equ:proof-bellman-discounted-return-3} and \ref{equ:state-action-relation}, the action-value function can be written as following

\begin{equation}
q_{\pi}(s, a)=\mathcal{R}_{s}^{a}+\gamma \sum_{s^{\prime} \in \mathcal{S}} \mathcal{P}_{s s^{\prime}}^{a} \sum_{a^{\prime} \in \mathcal{A}} \pi\left(a^{\prime} | s^{\prime}\right) q_{\pi}\left(s^{\prime}, a^{\prime}\right)
\label{equ:bellman-equation-action}
\end{equation}

It is easy to prove that there is always an optimal policy for any Markov decision process and it can be found by maximizing action-value function.

\begin{equation}
\pi_{*}(a | s)=\left\{\begin{array}{ll}{1} & {\text { if } a=\underset{a \in \mathcal{A}}{\operatorname{argmax}} q_{*}(s, a)} \\ {0} & {\text { otherwise }}\end{array}\right.
\label{equ:optimal-policy}
\end{equation}

Considering \ref{equ:optimal-policy}, Bellman optimality equation for action-value function can be obtained by replacing the policy in Equation \ref{equ:bellman-equation-action} with optimal policy. There are many ways to solve this equation such as Sarsa and Q-learning. This will be discussed in Section \ref{sec:td-learning}.

\subsection{Exploitation vs Exploration}\label{sec:explore}

If the agent has complete knowledge of the environment, in the other word, the transition probability $P_{ss^\prime}^a$ can be calculated given state $s$ and action $a$, Equation \ref{equ:bellman-optimality-action} can be solved by an iterative method with appropriate $\gamma$. However, this method is unable to deal with an unknown environment because a large amount of information has to be collected to estimate $P_{ss^\prime}^a$ before the convergence of action value function. If the $q$ function tends to be stable before the environment has been fully explored, the performance of the model would be far from satisfactory, especially in high action space situation.

To deal with this problem, $\epsilon$ - greedy selection \cite{sutton2018reinforcement} is introduced to ensure the agent make enough exploration before the convergence of the action value function. Instead of choosing the best action estimated by $q$ function, there is a probability of $\epsilon$ to randomly select from all actions. The mathematical expression of this method is shown as following

\begin{equation}
\pi(a | s)=\left\{\begin{array}{ll}{\epsilon / m+1-\epsilon} & {\text { if } a^{*}=\mathop{\arg\max}\limits_{a \in \mathcal{A}} q(s, a)} \\ {\epsilon / m} & {\text { otherwise }}\end{array}\right.
\label{equ:explor-exploit}
\end{equation}

where $m$ is the number of actions. This method may have a bad effect on the performance of the agent at first several episodes during the training but it can widen the horizon of the agent in long term view.

\subsection{Temporal Difference Learning}\label{sec:td-learning}

As mentioned in Section \ref{sec:explore}, most environment in the real world is unknown. To solve this problem, a method called Monte Carlo (MC) is used to sample data for estimating value function. The agent can learn the environment from one episode experience and the value function can be approximated by the mean of the return instead of the expectation. The mathematical expression can be described as following

\begin{equation}
v(S_t)=\frac{S(S_t)}{N(S_t)}
\label{equ:mc-update-value}
\end{equation}

where $S_t$ is the state at time $t$, $S(S_t)$ is the sum of return and $N(S_t)$ is the counter to record the visit number of state $S_t$. There are two kinds of visit: first visit and every visit. The former one means the model only need to record the first visit of state $S_t$ in one episode while all visit of $S_t$ in one episode will be taken into consideration in every visit. Simplify equation \ref{equ:mc-update-value}, we can get the recurrence equation for $v(s)$

\begin{equation}
v(s)\leftarrow v(s) + \eta(G_t-v(s))
\label{equ:mc-update}
\end{equation}

where $\eta$ is the learning rate which can control the update speed of the value function and $s$ is the state at time $t$. The problem of Monte Carlo method is all rewards in one episode have to be collected to get $G_t$. The value function can only be updated when reaching the end of the episode which may lead to low training efficiency. To update value function with an incomplete episode, the return can be replaced by estimated value function using bootstrapping. With the Bellman equation \ref{equ:bellman-expectation-state} and \ref{equ:mc-update}, we can write

\begin{equation}
v(s)\leftarrow v(s) + \eta(R_{t+1}+\gamma v(s^\prime)-v(s))
\label{equ:td-value}
\end{equation}

This idea is called Temporal Difference (TD) Learning. In TD learning, value function will be updated immediately after a new observation. Compared with MC methods, TD learning has lower variance because there are too many random actions $\{A_{t+1}, A_{t+2}, \cdots\}$ in the Monte Carlo method which will lead to the high variance. Similarly, the recurrence equation for action value function can be written as following

\begin{equation}
q(s,a)\leftarrow q(s,a) + \alpha\Big(R_{t+1}+\gamma q(s^\prime,a^\prime)-q(s,a)\Big)
\label{equ:td-action}
\end{equation}

where $s$ and $a$ is the state and action at time $t$, $s^\prime$ and $a^\prime$ is the state and action at time $t+1$. Equation \ref{equ:td-action} shows an iterative method to get the optimal action value function $q_*(s,a)$. With this equation and $\epsilon$ - greedy policy, the RL problem can be solved by Sarsa \cite{rummery1994line}.

\begin{algorithm}[ht]
\caption{Sarsa}
\begin{algorithmic}[1]
\State set learning rate $\alpha$, number of episodes $N$, explore rate $\epsilon$, discount factor $\gamma$
\State set $q(s,a)\gets 0$, $\forall s, a$
\For{episode $\gets 1$ to $N$}
    \State initialize time $t\gets 0$
    \State get state $s_0$ from the environment
    \State choose action $a_0$ following $\epsilon$ - greedy policy from $q(s,a)$
    \While{episode is incomplete}
        \State take action and get next state $s_{t+1}$, reward $r_{t+1}$ from the environment
        \State choose action $a_{t+1}$ following $\epsilon$ - greedy policy from $q(s,a)$
        \State update $q(s_t,a_t)\leftarrow q(s_t,a_t) + \alpha\Big(r_{t+1}+\gamma q(s_{t+1},a_{t+1})-q(s_t,a_t)\Big)$
        \State $t\gets t+1$, $s_{t}\gets s_{t+1}$, $a_{t}\gets a_{t+1}$
    \EndWhile
\EndFor
\end{algorithmic}
\label{alg:sarsa}
\end{algorithm}

The name of Sarsa is from the sequence $\{S_0,A_0,R_1,S_1,A_1,R_2,\cdots\}$. Besides Sarsa, there is another similar algorithm called Q learning \cite{watkins1989learning}.

\begin{algorithm}[ht]
\caption{Q learning}
\begin{algorithmic}[1]
\State set learning rate $\alpha$, number of episodes $N$, explore rate $\epsilon$, discount factor $\gamma$
\State set $q(s,a)\gets 0$, $\forall s, a$
\For{episode $\gets 1$ to $N$}
    \State initialize time $t\gets 0$
    \State get state $s_0$ from the environment
    \While{episode is incomplete}
        \State choose action $a_t$ following $\epsilon$ - greedy policy from $q(s,a)$
        \State take action and get next state $s_{t+1}$, reward $r_{t+1}$ from the environment
        \State update $q(s_t,a_t)\leftarrow q(s_t,a_t) + \alpha\Big(r_{t+1}+\gamma \max\limits_{a}q(s_{t+1},a)-q(s_t,a_t)\Big)$
        \State $t\gets t+1$, $s_{t}\gets s_{t+1}$
    \EndWhile
\EndFor
\end{algorithmic}
\label{alg:qlearning}
\end{algorithm}

\begin{figure}[ht]
\centering
\includegraphics[width=15cm]{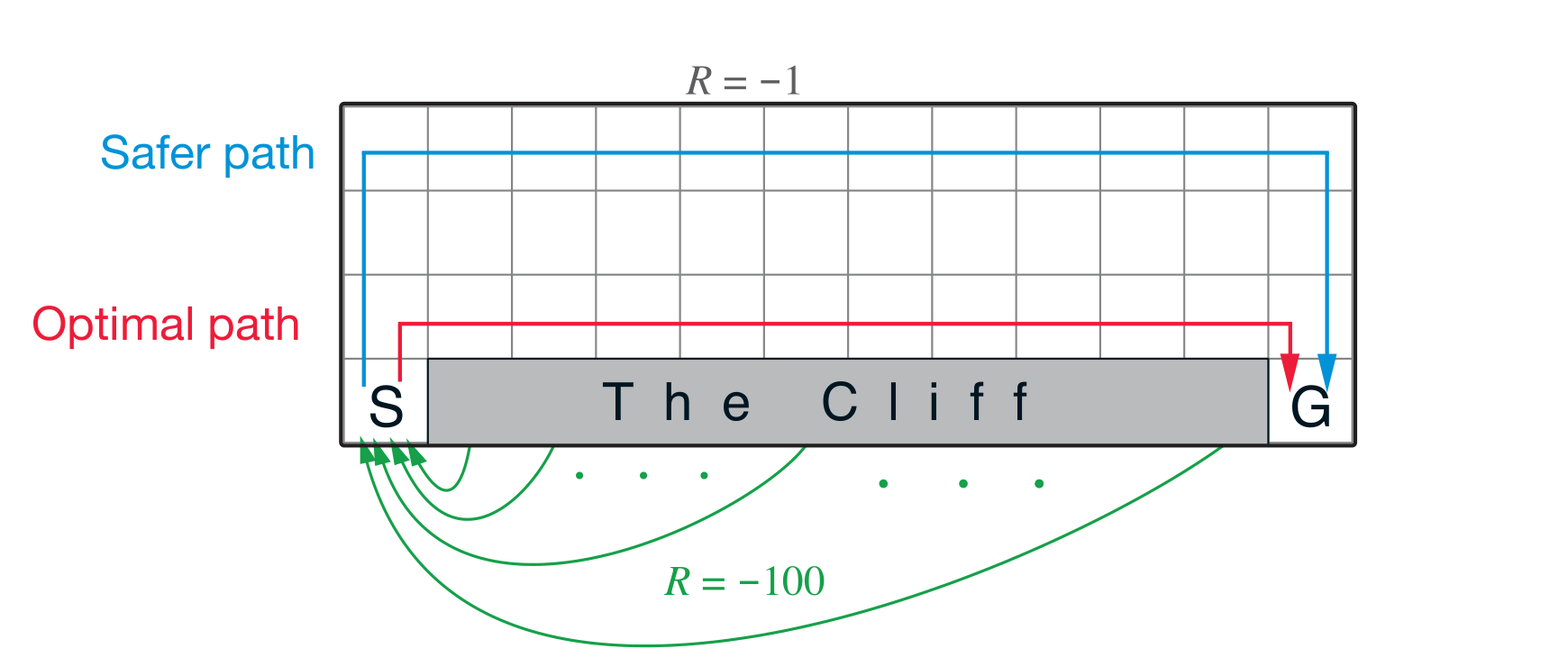}
\caption{Example of cliff walking from \cite{sutton2018reinforcement}.}
\label{fig:cliff-walking}
\end{figure}

In algorithm \ref{alg:qlearning}, there are two policies during the iteration. When choosing the action $a_{t+1}$ from $q(s,a)$ given $s_{t+1}$, Sarsa uses $\epsilon$ - greedy policy while Q learning uses greedy policy. But both of them are choosing $a_t$ with $\epsilon$ - greedy policy. Considering the example of Cliff Walking shown in Figure \ref{fig:cliff-walking} from Sutton's book \cite{sutton2018reinforcement}, every transition in the environment will get $-1$ reward except next state is the cliff which the agent will get $-100$ reward, Sarsa is more likely to choose the safe path while Q learning tends to choose the optimal path with $\epsilon$ - greedy policy. But both of them can reach the optimal policy if reducing the value of $\epsilon$.

\subsection{Deep Q Network}\label{sec:dqn}

Q learning is a powerful algorithm to solve simple reinforcement problems. However, it is unable to deal with continuous states or continuous actions. To solve the former problem, deep learning method can be used to approximate action value function.

Generally, states are image data observed by the agent and convolutional neural network is an effective way to extract features from this kind of data in convolution layers and feed them into the fully connected layer to approximate $q$ function. Several consistent stationary images will be stacked into one input data to make the model understand that the agent is moving. But the input data is highly dependent, the performance of the model will be largely affected by the dependency.

As mentioned in Section \ref{sec:rl}, DeepMind introduced experience replay pool which will store the experience into the memory and sample some of them to optimize the neural network model in 2013 \cite{mnih2013playing}. Using Q learning, deep learning and experience replay pool, the improved algorithm named Deep Q Network (DQN) shows incredible performance on Atari games according to their paper. Two years later, they found the agent became more stable by using two network \cite{mnih2015human}. This algorithm can be described as below

\begin{algorithm}[ht]
\caption{Deep Q Network}
\begin{algorithmic}[1]
\State initialize policy network $q(s,a)$ with random weights
\State set learning rate $\alpha$, number of episodes $N$, explore rate $\epsilon$, discount factor $\gamma$
\State set batch size $M$, update step $T$
\State set target network $q^\prime(s,a)=q(s,a)$
\For{episode $\gets 1$ to $N$}
    \State initialize time $t\gets 0$
    \State get state $s_0$ from the environment
    \While{episode is incomplete}
        \State choose action $a_t$ following $\epsilon$ - greedy policy from policy network $q(s,a)$
        \State take action and get next state $s_{t+1}$, reward $r_{t+1}$ from the environment
        \State store transition $(s_t,a_t,s_{t+1},r_{t+1})$ in experience replay pool
        \State random sample $M$ batch experience $(s_k,a_k,s_{k+1},r_{k+1})$ from the pool
        \State calculate corresponding $q(s_k,a_k)$ from policy network $q(s,a)$
        \State calculate $y_k$ using target network $q^\prime(s,a)$
        \begin{equation*}
        y_k=\left\{\begin{array}{ll}{r_{k+1}} & {\text { if next state is completed }} \\ {r_{k+1} + \gamma\max\limits_{a}q^\prime(s_{k+1}, a)} & {\text { otherwise }}\end{array}\right.
        \end{equation*}
        \State optimize the policy model with gradient $(y_k-q(s_k,a_k))^2$
        \State replace target network with policy network when reach the update step $T$
        \State $t\gets t+1$, $s_{t}\gets s_{t+1}$
    \EndWhile
\EndFor
\end{algorithmic}
\label{alg:dqn}
\end{algorithm}

All states in Algorithm \ref{alg:dqn} have to be pre-processed before feeding into a neural network model. Based on Deep Q Network, there are three kinds of improved algorithms considering the stability of the training process, the importance of each experience and new neural network architecture. Double DQN \cite{van2016deep} utilizes the advantage of two networks. Instead of finding the optimal $q$ value from target network $q^\prime(s,a)$ directly, this method chooses the optimal action from the policy network and find the corresponding $q$ value in the target network. Use the term in Algorithm \ref{alg:dqn}, the change can be illustrated as following

\begin{equation}
y_k=\left\{\begin{array}{ll}{r_{k+1}} & {\text { if next state is completed }} \\ {r_{k+1} + \gamma q^\prime(s_{k+1}, \mathop{\arg\max}\limits_{a}q(s_{k+1}, a))} & {\text { otherwise }}\end{array}\right.
\label{equ:double-dqn-cal}
\end{equation}

Prioritized Experience Replay (PER) introduced a way to efficiently sample transitions from the experience replay pool \cite{schaul2015prioritized}. Instead of uniform random sampling, there is a priority of each transition

\begin{equation}
P(i)=\frac{p_{i}^{\alpha}}{\sum_{k} p_{k}^{\alpha}}
\label{equ:priority-distribution}
\end{equation}

where $p_i>0$ is the priority of transition $i$ and $\alpha$ is the indicator of the priority, especially $\alpha=0$ when using uniform random sampling. The priority can be measured by TD error $\delta_i$, which is the following term

\begin{equation}
\delta_i=R_i+\gamma \max\limits_{a} q(s_i, a)-q(s_{i-1},a_{i-1})
\end{equation}

Based on TD error, $p(i)$ can be calculated in two way. The first is proportional prioritization which uses the absolute value of TD error 

\begin{equation}
p(i)=|\delta|+\epsilon
\end{equation}

where $\epsilon$ is to avoid zero prioritization. The other one is rank-based

\begin{equation}
p(i)=\frac{1}{\text{rank}(i)}
\end{equation}

where $\text{rank}(i)$ is the rank of transition by sorting TD error $\delta_i$. According to Schaul, both proportional based and rank based prioritization can speed-up the training but the later one is more robust which has better performance when meeting outliers.

However, the random sampling is abandoned after adding priority mechanism which will result in high bias. In other words, those transitions with small TD error are unlikely to be sampled and the distribution is changed. Therefore, the final model may far from the optimal policy and performance of the agent even be lower than DQN. Important sampling (IS) \cite{neal2001annealed} is an effective technique to estimate a distribution by sample data from a different distribution. Given a probability density function $p(x)$ over distribution $\mathcal{D}$, with the definition of the expectation

\begin{equation}
\mathbb{E}_p\left[f(x)\right]=\int_{\mathcal{D}} f(x) p(x) \mathrm{d} x
\end{equation}

where $\mathbb{E}_p\left[\cdot\right]$ denotes the expectation for $x\sim p$ and $f$ is the integrand. Given another probability density function $q(x)$, the expectation can be written as following

\begin{equation}
\int_{\mathcal{D}} f(x) p(x) \mathrm{d} x=\int_{\mathcal{D}} \frac{f(x) p(x)}{q(x)} q(x) \mathrm{d} x=\mathbb{E}_{q}\left[\frac{f(x) p(x)}{q(x)}\right]
\label{equ:importance-sampling}
\end{equation}

where $\mathbb{E}_q\left[\cdot\right]$ denotes the expectation for $x\sim q$. With Monte Carlo integration, the expectation $\mathbb{E}_p\left[f(x)\right]$ can be estimated by

\begin{equation}
\frac{1}{N}\sum_{i=1}^N\frac{f(i)p(i)}{q(i)}
\end{equation}

where $i$ is sampled from $x$. if $p(i)$ is uniform distribution and $q(i)$ refers to Equation \ref{equ:priority-distribution}, we have

\begin{equation}
\frac{p(i)}{q(i)}=\frac{1}{N}\cdot\frac{1}{P(i)}
\end{equation}

adding a tunable parameter $\beta$, we obtain the importance-sampling weights

\begin{equation}
w_{i}=\frac{\left(N \cdot P(i)\right)^{-\beta}}{\max\limits_{i}w_i}
\end{equation}

where $\beta$ will decay from a user-defined initial value to $1$ and the bias completely disappears when $\beta=1$. Term $\max\limits_{i}w_i$ is used to normalize the weight to increase stability. Use the term in Algorithm \ref{alg:dqn}, the update of $q$ function can be modified as following

\begin{equation}
q(s,a) \leftarrow q(s,a) + \eta \cdot w_{k} \cdot \nabla (y_k-q(s_k,a_k))^2
\end{equation}

where $\delta_{k}$ is TD error and $\eta$ is learning rate.

Dueling DQN architecture used a new concept called advantage function which is the subtraction of the action value function and state value function \cite{wang2015dueling}.

\begin{equation}
A_{\pi}(s,a)=q_{\pi}(s,a)-v_{\pi}(s)
\label{equ:advantage-function}
\end{equation}

\begin{figure}[ht]
\centering
\includegraphics[width=15cm]{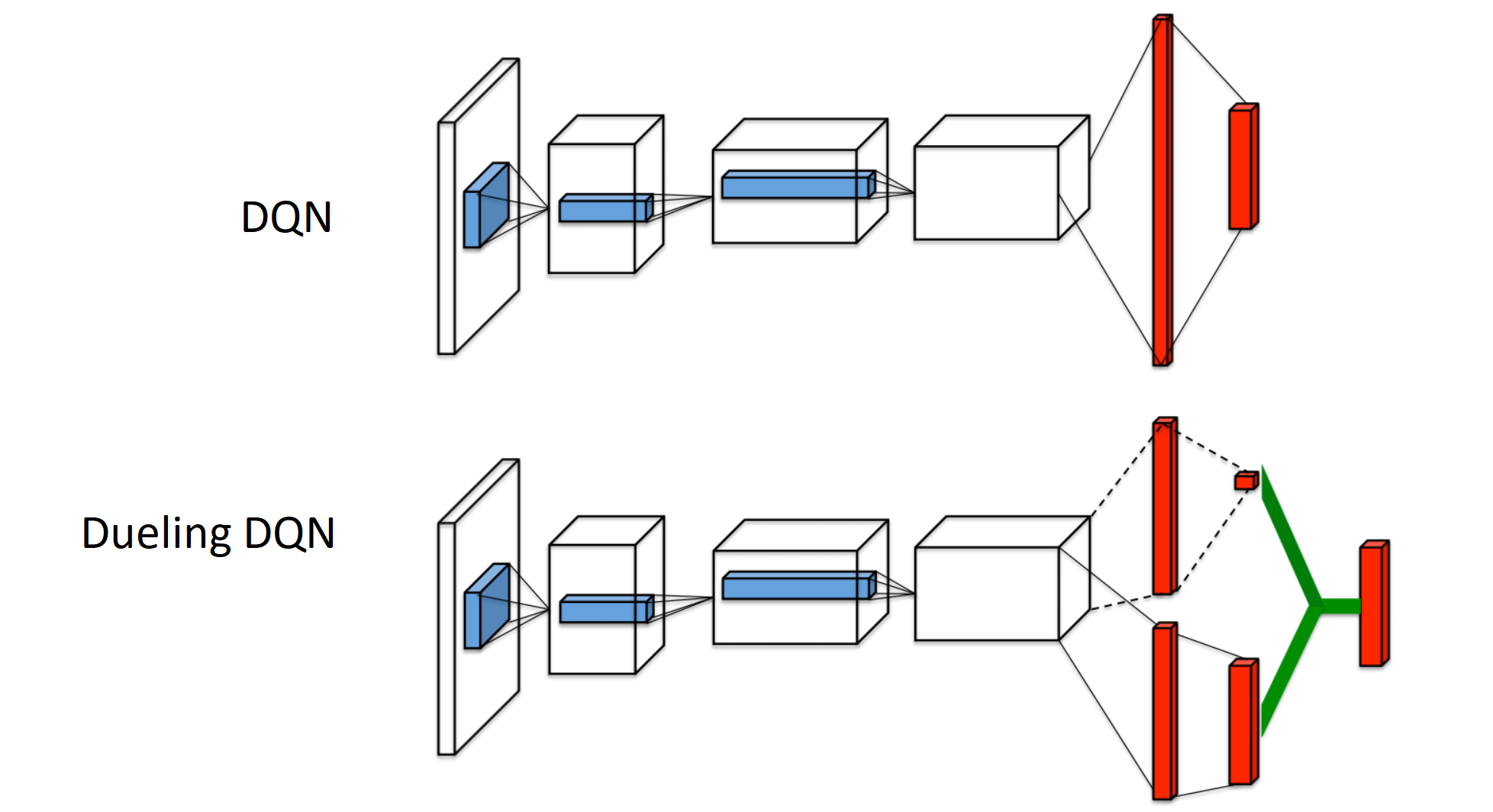}
\caption{Architecture comparison between Dueling DQN and DQN \cite{wang2015dueling}}
\label{fig:duel-dqn}
\end{figure}

As shown in Figure \ref{fig:duel-dqn}, dueling network architecture use summation of two steams which is advantage function and state value function to get the $q$ function. The state values can be updated more accurately with this method.

\chapter{Methodology}

This chapter gives the requirements of the project, introduces the design of reward function and shows the preprocessing steps of the input image. The choice of the model, as well as the architecture, will be discussed. Experiment design and evaluation methods will be illustrated in the last.

\section{Requirements}

\subsection{Software Requirement}

Considering the readability of the code, widely used additional frameworks such as Torch, Python is a suitable choice for this project. OpenCV is used to preprocess the image getting from the environment. Numpy is a Python library which accelerates matrices operations with C. This enables the user to write efficient scientific computing code with Python. There are plenty of deep learning frameworks like Tensorflow which has many extensive API and is widely used in industrial products. However, it will take a relatively long time for the beginner to fully understand the usage of Tensorflow. Pytorch is a recently developed framework which is described as "Numpy with GPU". The simplicity of Pytorch makes more and more academic researchers using it to implement their new ideas in a much easier way. Because T-rex Runner is running on Chrome, the latest Chrome is used here. Gym is a game library developed by OpenAI \cite{brockman2016openai}. This framework provides a built-in environment for some famous games such as Atari 2600 and it is easy for the user to customize their own environment. Table \ref{tab:software-requirement} shows all software requirement in this project.

\begin{table}[ht]
\centering
\begin{tabular}{ll}
\hline\hline
\textbf{Software} & \textbf{Description} \\
\hline
OS & Windows 10 \\
Programming language & Python 3.7.4 \\
Framework & OpenCV, Pytorch, Numpy, Gym \\
Browser & Chrome 76 \\
\hline
\end{tabular}
\caption{Software requirement}
\label{tab:software-requirement}
\end{table}

\subsection{Hardware Requirement}

As the game is running on Chrome, it is hard to use a Linux server to perform the experiments. Although headless Chrome is a plausible choice, there are some environmental issues during the investigation. Therefore, all experiments will be running on the laptop from the author. There will be some limitation such as 6GB GPU memory limits the size of experience replay pool. Therefore, parameters related to hardware limitation will be suitably chosen without tuning in this project. Table \ref{tab:hardware-requirement} lists all hardware information used in this project.

\begin{table}[ht]
\centering
\begin{tabular}{ll}
\hline\hline
\textbf{Hardware} & \textbf{Description} \\
\hline
CPU & Intel Core i5-8300H \\
RAM & 16G \\
GPU & Nvidia GTX 1060 6G \\
\hline
\end{tabular}
\caption{Hardware requirement}
\label{tab:hardware-requirement}
\end{table}

\section{Game Description}\label{sec:game-description}

\begin{figure}[ht]
\centering
\includegraphics[width=13cm]{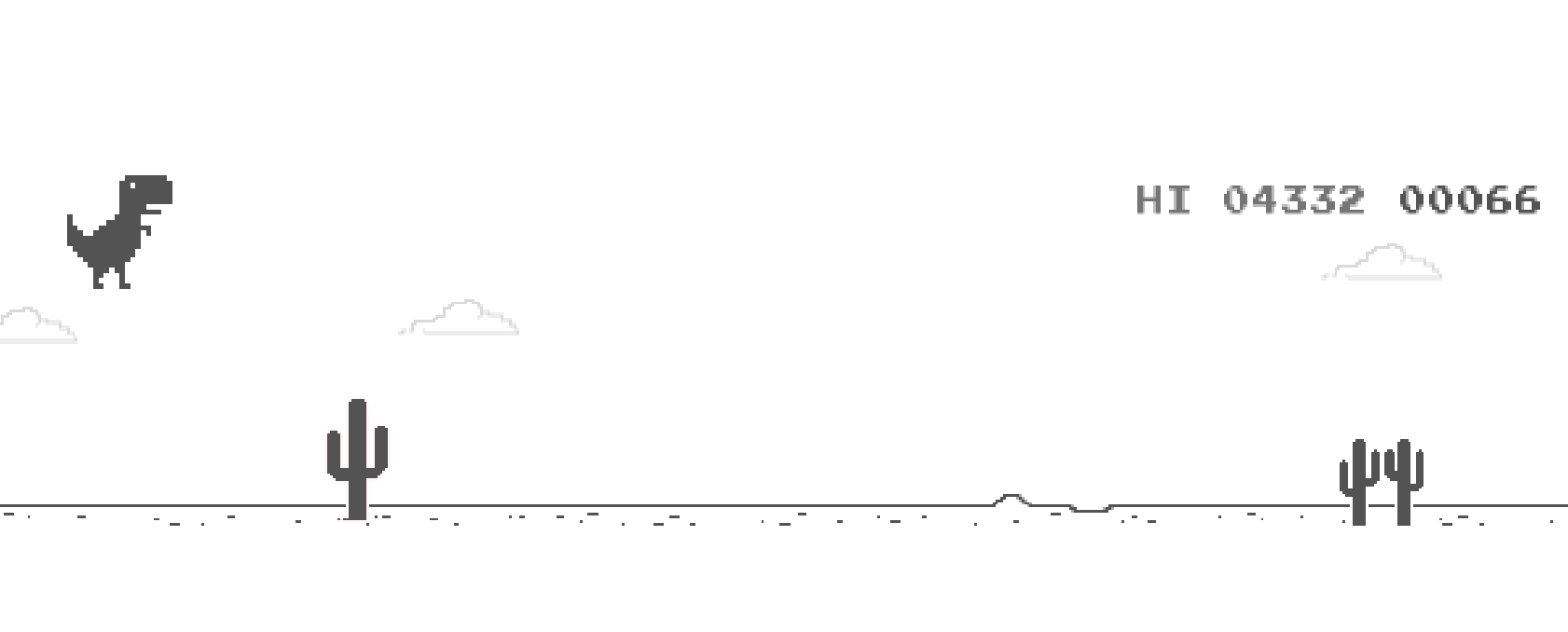}
\caption{A screenshot of T-rex Runner.}
\label{fig:dino-run}
\end{figure}

\begin{figure}[ht]
\centering
\begin{subfigure}{.3\textwidth}
  \includegraphics[width=.9\linewidth]{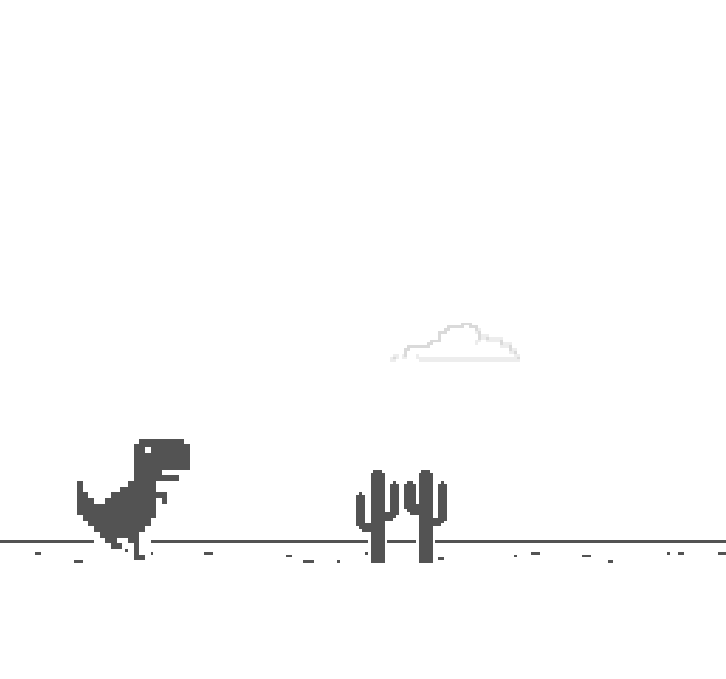}
  \caption{Do nothing}
  \label{fig:dino-do-nothing}
\end{subfigure}
\begin{subfigure}{.3\textwidth}
  \includegraphics[width=.9\linewidth]{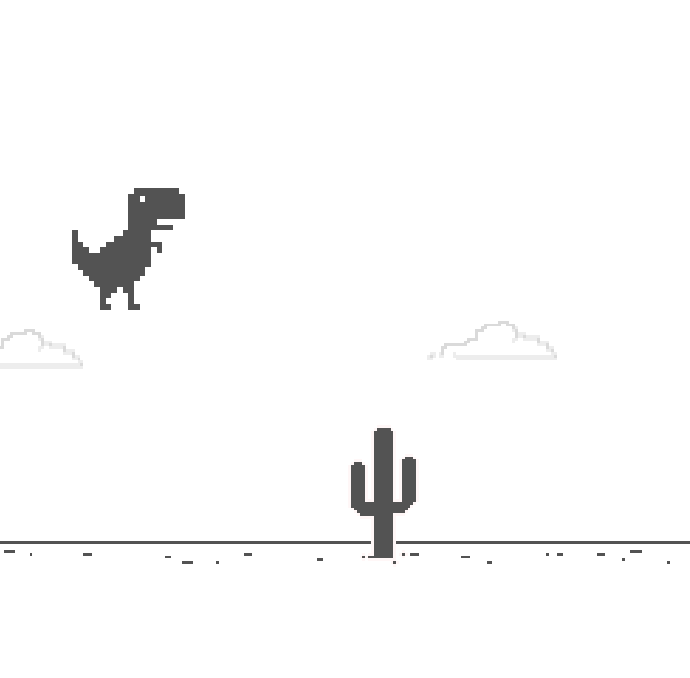}
  \caption{Jump}
  \label{fig:dino-jump}
\end{subfigure}
\begin{subfigure}{.3\textwidth}
  \includegraphics[width=.9\linewidth]{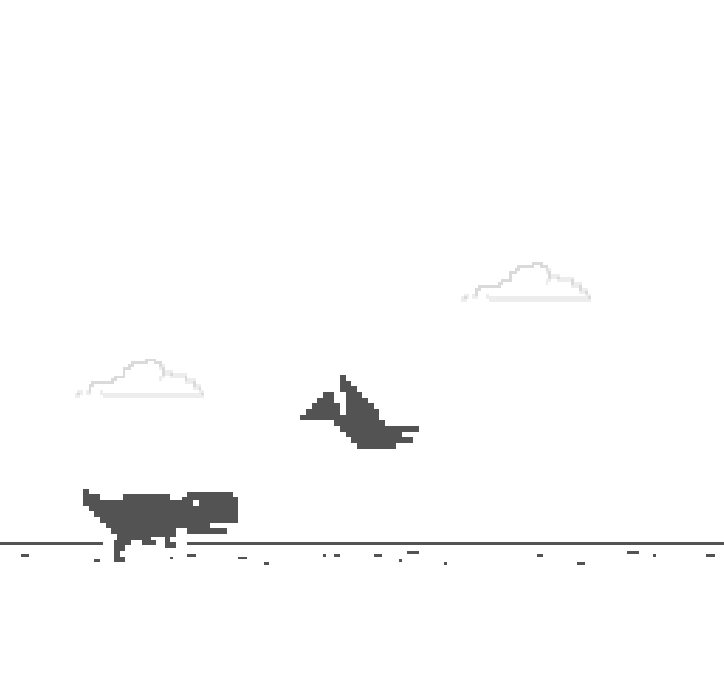}
  \caption{Duck}
  \label{fig:dino-duck}
\end{subfigure}
\caption{Three type of actions in T-rex Runner}
\label{fig:dino-action}
\end{figure}

T-rex Runner is a dinosaur game from Google Chrome offline mode. Everyone can access \href{http://wayou.github.io/t-rex-runner/}{this link} on Chrome to play the game. The target for players is to control the dinosaur overcoming as many obstacles as possible. The current score of the game will increase by time if the dinosaur keeps alive as shown at the top right corner of Figure \ref{fig:dino-run} as well as the highest score. As shown in Figure \ref{fig:dino-action}, the dinosaur has three actions to choose in every state: do nothing, jump or duck.

Environment plays an important role in reinforcement learning because the agent will improve the policy based on the feedback from it. However, it is difficult to quantify the rewards for each action as well as the return for an entire episode. In most research for RL algorithms, modifying reward will not be taken into consideration but it will significantly impact the performance of the model because it decides the behavior of the agent. For example, shaping reward shows a better performance in Andrew's experiment\cite{ng1999policy}. It adds a new term $F$ to modify the original reward based on the goal

\begin{equation}
R^{\prime}=R+F
\end{equation}

The closer the agent towards the goal, the larger the $F$ is. However, the aim of this project is to train the agent to play the game and compare the performance between different algorithms. So the effect of reward function will not be taken into consideration and a fixed reward function will be used across all experiments.

Since there is no previous study on T-rex Runner with reinforcement learning, the design of reward function is a hard part of this project. Intuitively, the best design is awarding the agent for jumping over the obstacles and penalizing it for hitting the obstacles. The jumping reward will gradually increase as time goes by. However, object detection in moving pictures is required to fulfill this goal. As this task is out of the requirements of this project, we proposed a naive reward design as shown in Algorithm \ref{alg:reward-design}.

\begin{algorithm}[ht]
\caption{Reward Design in T-rex Runner}
\begin{algorithmic}[1]
\If{episode is completed}
    \State return reward as $-1$
\Else
    \If{agent choose jump}
        \State return reward as $0$
    \Else
        \State return reward as $0.1$
    \EndIf
\EndIf
\end{algorithmic}
\label{alg:reward-design}
\end{algorithm}

The basic idea of Algorithm \ref{alg:reward-design} is giving a relatively small reward to the agent if it is alive and penalize it when hitting an obstacle. Zero reward for jumping is set to make the dinosaur only jumps if it is very close to obstacles. The unexpected jump will limit the movement in the next few states. 

Although there are three kinds of action in this game as introduced in Section \ref{sec:game-description}, duck is optional because the agent can overcome the obstacle using jump under the same circumstances. Considering most obstacles in the game are cactus which can only be overcome by jumping, only two actions (do nothing and jump) will be used in this investigation.

\section{Model Selection}

Since there are only two actions in T-rex Runner, according to the literature review on deep reinforcement learning in Section \ref{sec:rl}, value-based methods are proved to be powerful to handle this game. Although policy-based methods such as proximal policy gradient is a good choice too, only DQN, double DQN, DQN with prioritized experience replay and dueling DQN will be investigated in this project due to the time limitation.

Deep Q network which is shown in Algorithm \ref{alg:dqn} is a basic reinforcement learning algorithm using deep learning. According to the result from DeepMind, it is expected to achieve at least human-level results with only DQN.

Double DQN mitigates the $q$ value overestimation problems utilizing two advantage of two networks as shown in Equation \ref{equ:double-dqn-cal} but it is not expected to achieve a higher performance in this experiment because there is only two actions. The bad effect of overestimation problems is not obvious under this circumstance.

Dueling DQN adds an advantage function which is the subtraction of action value function and state value function before the output layer in the convolutional neural network as shown in Equation \ref{equ:advantage-function}. Since the evaluated game in \cite{wang2015dueling} is a similar racing game overcoming obstacles compared with T-rex Runner, this algorithm is expected to have a better performance than DQN.

Prioritized Experience replay improves training efficiency by changing the distribution of the stored transitions. It assigns the weight for each experience by TD error. There are two ways to calculate prioritization which is proportional based method and rank-based method. According to the \cite{schaul2015prioritized}, the former one has a relatively better performance, only this method will be implemented in this investigation due to the time limitation. The performance is expected to be the same as DQN because there is no change in the algorithm but it may be faster to reach the same performance.

\section{Image Preprocessing}

\begin{figure}[ht]
\centering
\includegraphics[width=10cm]{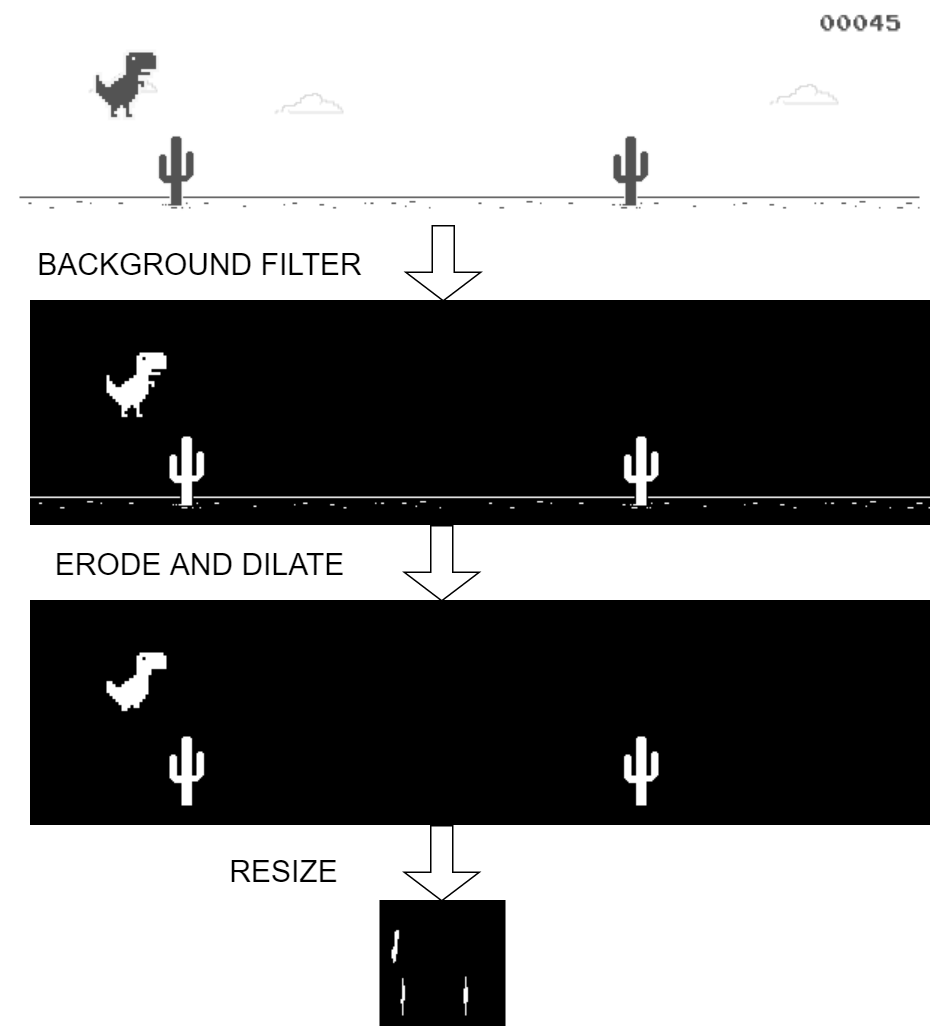}
\caption{Preprocessing steps for T-rex Runner.}
\label{fig:dino-preprocess}
\end{figure}

Following the preprocessing step in \cite{mnih2013playing, mnih2015human}, the raw observed image which is in RGB representation will be converted to gray-scale representation. To make the network easier to recognize dinosaur and obstacles, unnecessary objects such as clouds and scores will be removed. In this step, the color of the background and the object are reversed in order to perform erosion and dilation. These two basic morphological operations can help reduce small bright color which is often noisy data. Finally, the image is resized to $84\times 84$ following the recipe from DeepMind. Since the movement should be recognized by the neural network, perform the same preprocessing step for last four frames in the history and stack those four as one data point which is also the input of CNN. The entire process is shown in Figure \ref{fig:dino-preprocess}.

\section{Convolutional Neural Network Architecture}

\begin{figure}[ht]
\centering
\includegraphics[width=13cm]{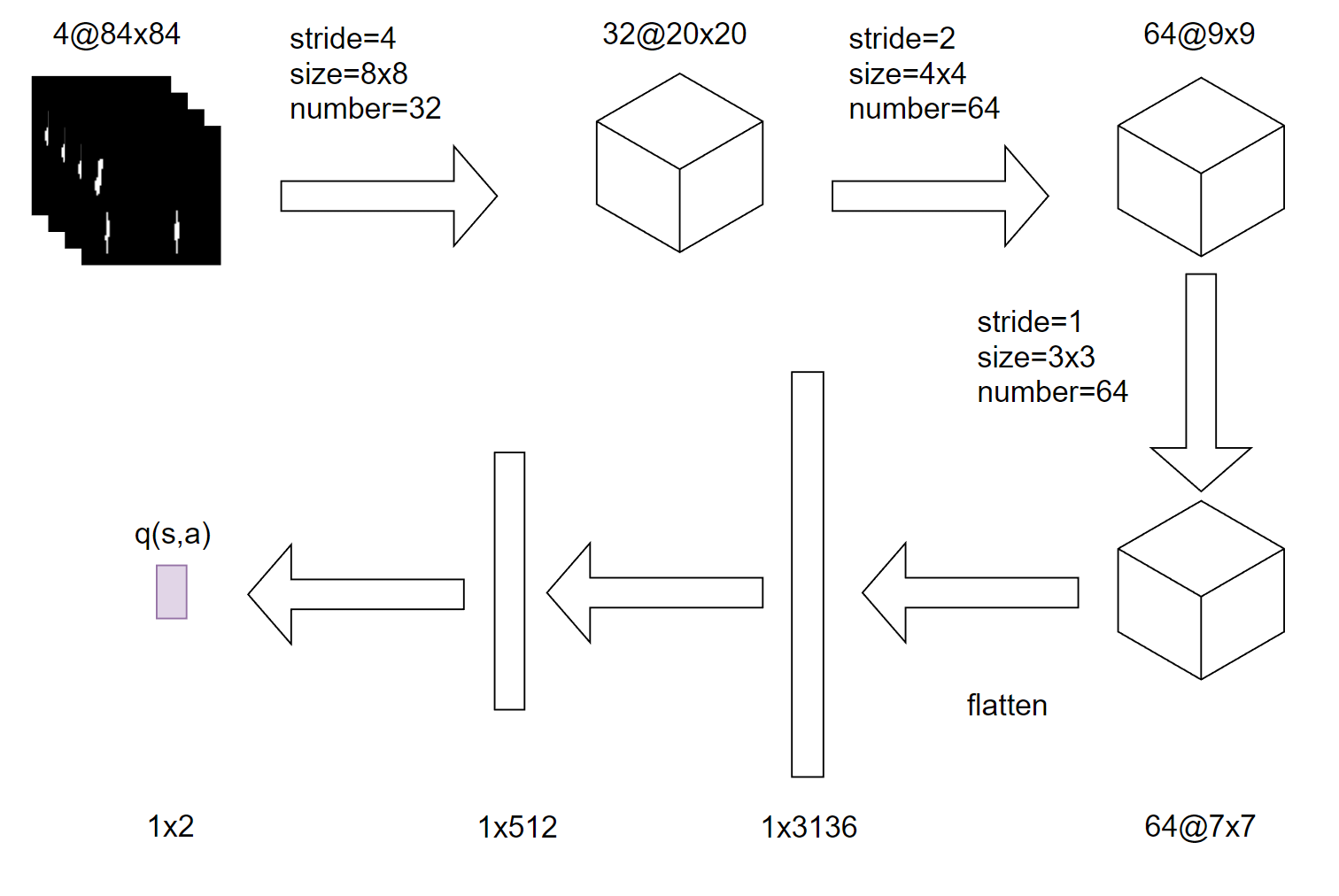}
\caption{Convolutional Neural Network architecture for Deep Q Network.}
\label{fig:cnn-architecture}
\end{figure}

There are two kinds of convolutional neural network used in this project. The basic DQN is proposed in \cite{mnih2013playing, mnih2015human} which used three convolutional layers and two fully connected layers. The reason for not using pooling layer is to detect the movement of the agent. Both max pooling and average pooling may make the neural network ignore a very small change in the image. Therefore, there are only convolutional layers in this architecture. The architecture for training the agent using DQN is shown in Figure \ref{fig:cnn-architecture}. 

\begin{figure}[ht]
\centering
\includegraphics[width=13cm]{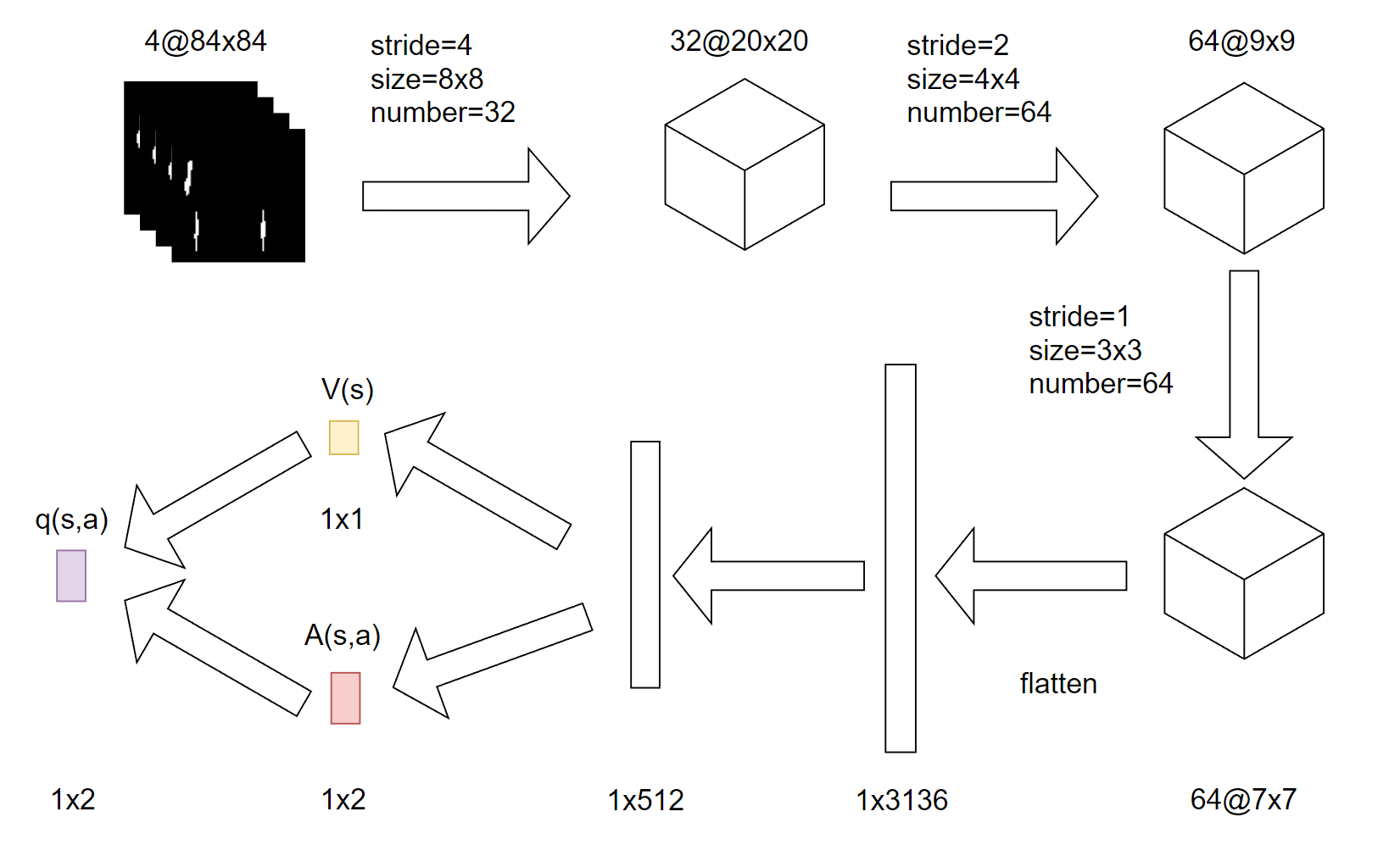}
\caption{Convolutional Neural Network architecture for Dueling Deep Q Network.}
\label{fig:duel-cnn-architecture}
\end{figure}

Dueling architecture is proposed in \cite{wang2015dueling} which divided the $q$ network into two parts. One of them is only related to the state value function $v(s)$, the other one is advantage function $A(s,a)$ which is affected by both state and action. The final action value function is the summation of those two.

\begin{equation}
q(s,a;\theta,\omega_1,\omega_2)=v(s;\theta,\omega_1)+A(s,a;\theta,\omega_2)
\end{equation}

where $\theta$ is the shared parameter of CNN, $\omega_1$ is the value function only parameter and $\omega_2$ is the advantage function only parameter. Both DQN and Dueling DQN are using Algorithm \ref{alg:dqn}, the only difference is the neural network architecture. RMSprop \cite{tieleman2012lecture} which is an adaptive gradient method based on stochastic gradient descent will be used as the optimization algorithm in this project. This is the same optimization method used by DeepMind \cite{mnih2013playing, mnih2015human}. Figure \ref{fig:duel-cnn-architecture} shows the process of dueling DQN.

\section{Experiments}

\subsection{Hyperparameter Tuning}\label{sec:design-hyper-parameter-tuning}

Before the comparison of algorithms, hyperparameter tuning is required to get high-performance models. As mentioned before, the memory size is fixed to $3\times 10^5$ due to the hardware limitation. Because there is no previous study on this game, and the hyperparameters list in \cite{mnih2015human} have a bad result on this game. All other hyperparameters have to be set to a suitable value. Grid search is performed to find a workable combination of those parameters.

Due to the time limitation, all parameters will only be slightly modified and only one hyperparameter will vary during each tuning experiment. The choice of the parameter will consider both score and stability. Each parameter will be tuned with $800$ episodes.

\subsection{Comparison of different Deep Q Network Algorithms}\label{sec:design-dqn-compare}

There are three improved reinforcement algorithms based on DQN mentioned in Section \ref{sec:dqn}. Double DQN makes the performance of the agent more stable by solving the overestimated $q$ value problem. Prioritized experience replay improves the training efficiency by sample more valuable transitions. Dueling DQN modifies the neural network architecture to get a better estimation of state values. 

In this experiment, DQN will be first used to train the agent based on the hyperparameters tuned in Section \ref{sec:design-hyper-parameter-tuning} and this result will be treated as a baseline across all the experiments. Double DQN, DQN with prioritized experience replay and Dueling DQN will be applied to the agent separately. The performance of those three is expected to be better than DQN according to the related papers. Due to time limitation, no combination of those three algorithms will be performed in this project. This section only compares the performance of each algorithm.

\subsection{Effect of Batch Normalization}

As mentioned in Section \ref{sec:bn}, it is proved that batch normalization can reduce training time and mitigate the vanishing gradient problem in a convolutional neural network. However, there is no evidence that this method has the same effect on reinforcement learning. This section will perform experiments on this point. Based on the experiment in Section \ref{sec:design-dqn-compare}, adding batch normalization in each convolutional layer and compared with the results with the outcome in previous experiments.

\section{Evaluation}

To evaluate the performance of the agent, DeepMind used trained agent playing the game for $30$ times for up to $5$ min and $\epsilon$ - greedy policy with $\epsilon=0.05$ \cite{mnih2015human}. Considering only one game is investigated in this project, the average score will be used instead of average reward because the number of jumps in each episode will affect the total reward according to the designed reward function. The greedy policy will be used in the evaluation stage instead of $\epsilon$ - greedy policy because the later one will bring randomness to the decision which will affect the performance of the trained model. Therefore, the trained agent will play the game for $30$ times without time limitation and using greedy policy. All outcomes will be compared with the results from a human expert.

The average scores during the training stage will be shown graphically. This is a clear way to show the learning efficiency of each algorithm. Both graphical and statistical results such as mean, variance and median will be analyzed. However, only statistical results will be analyzed in the testing stage because the trained model for each algorithm are the same and there is no increasing trend can be shown like in the training stage. These results will be visualized with a boxplot.

\chapter{Results and Discussion}

\section{Hyper Parameter Tuning}

The value of hyperparameters may affect the performance of the model. However, there are so many parameters in reinforcement learning including optimization algorithm parameters such as learning rate. This may take a long time to find the optimal combination of these parameters using a grid search. Since there is no metric like accuracy in RL which can easily reflect the performance of the model, we assume each parameter is independent of others. Therefore, each parameter can be tuned one after another. Because the objective of this project is to compare the performance between different algorithms and the effect of batch normalization, those tuned parameters by DQN will be used across all the experiments. The start hyperparameters of DQN are shown in Table \ref{tab:first-param-info}.

\begin{table}[ht]
\centering
\begin{tabular}{llp{8cm}}
\hline\hline
\textbf{Hyper parameter} & \textbf{Value} & \textbf{Description} \\
\hline
Memory Size & $3\times 10^5$ & Size of experience replay pool \\
Batch Size & 128 & Size of minibatch to optimize model \\
Gamma & 0.99 & Discount factor \\
Initial $\epsilon_0$ & $1\times 10^{-1}$ & Explore probability at the start of the training \\
Final $\epsilon^\prime$ & $1\times 10^{-3}$ & End point of explore probability in $\epsilon$ decay \\
Explore steps & $1\times 10^5$ & Number of steps for $\epsilon$ decay from $\epsilon_0$ to $\epsilon^\prime$ \\
Learning Rate & $1\times 10^{-4}$ & Learning speed of the model \\
\hline
\end{tabular}
\caption{Hyper parameters used in all experiments}
\label{tab:first-param-info}
\end{table}

\subsection{Learning Rate}

Learning rate controls the learning speed of the model, too large value will result in divergence and too small value may double the training time. 

\begin{figure}[ht]
\centering
\includegraphics[width=13cm]{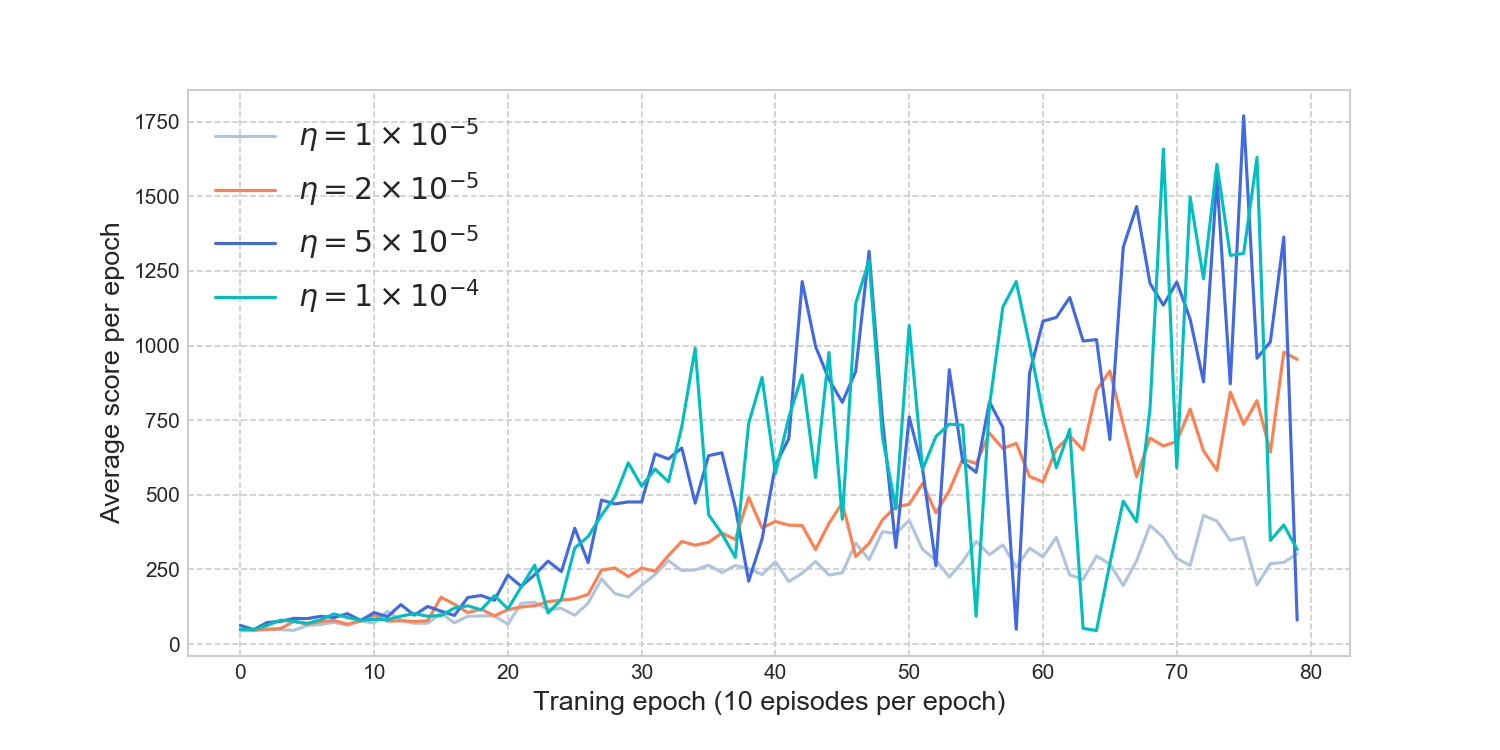}
\caption{Hyper parameter tuning for learning rate.}
\label{fig:tune-lr}
\end{figure}

Figure \ref{fig:tune-lr} shows four different values of learning rate. Obviously, $1\times 10^{-5}$ is too small and there is no increase trend during the entire process. Both $1\times 10^{-4}$ and $5\times 10^{-5}$ make the score unstable after 50th epoch. Considering the stability and $200$ epochs will be trained in formal experiment, $2\times 10^{-5}$ will be chosen as learning rate.

\subsection{Batch Size}

Batch size defines how many transitions will be used to update the neural network which may affect the training speed. But as mentioned in \ref{sec:rl}, too big size will cause the dependency problems which may largely affect the performance of the model.

\begin{figure}[ht]
\centering
\includegraphics[width=13cm]{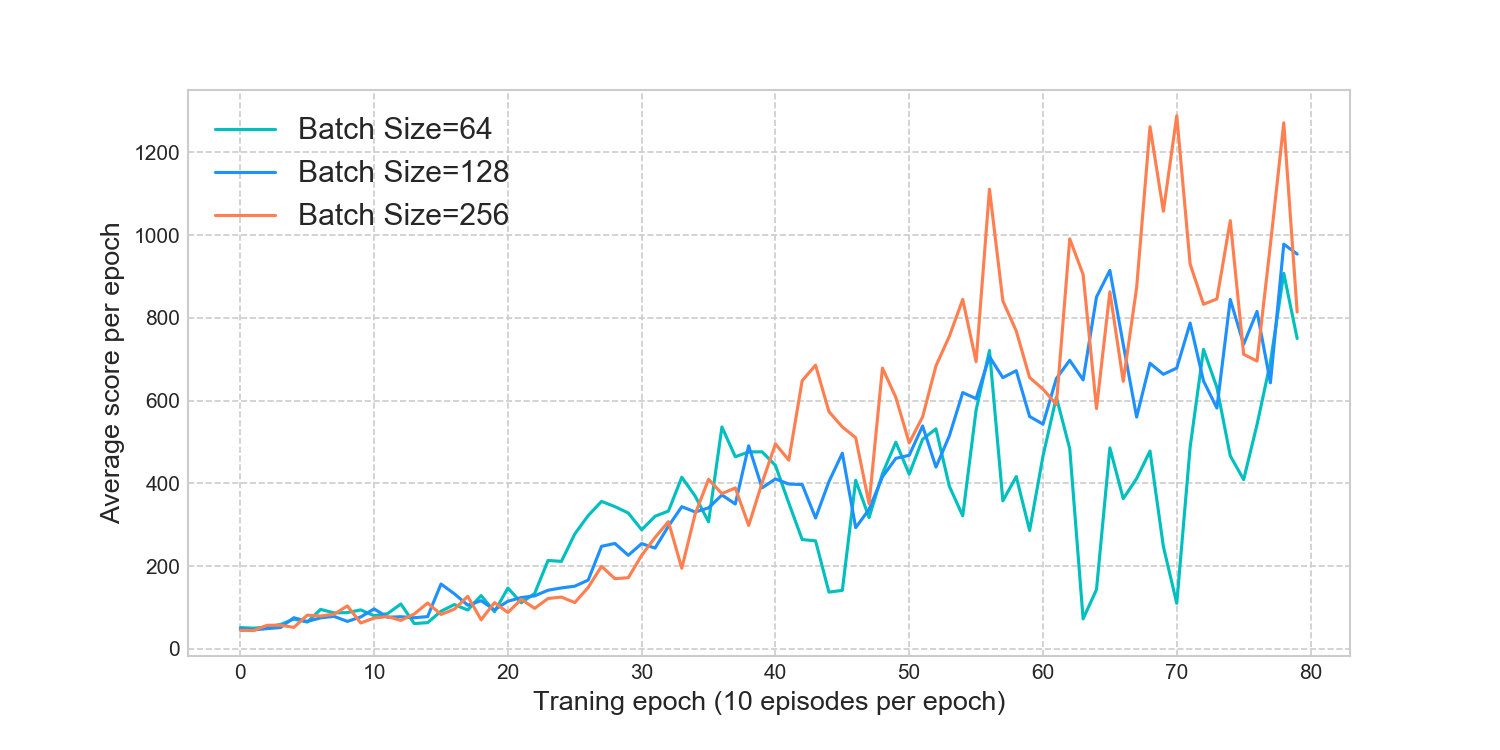}
\caption{Hyper parameter tuning for batch size.}
\label{fig:tune-batchsize}
\end{figure}

As shown in Figure \ref{fig:tune-batchsize}, the average score of three curves at epoch $80$ are all around $800$. Among those three, the most stable one is batch size $128$.

\subsection{Epsilon}

$\epsilon$ - greedy policy determines the probability of exploration. In some games, especially with high action spaces, this value can affect how good the model will converge. However, there are only two actions in T-rex Runner so it is unnecessary to random choose action at the begin. Instead of initializing $\epsilon$ to $1$ as DeepMind did in their paper \cite{mnih2015human}, the start value is set to $0.1$ in this model.

\begin{figure}[ht]
\centering
\includegraphics[width=13cm]{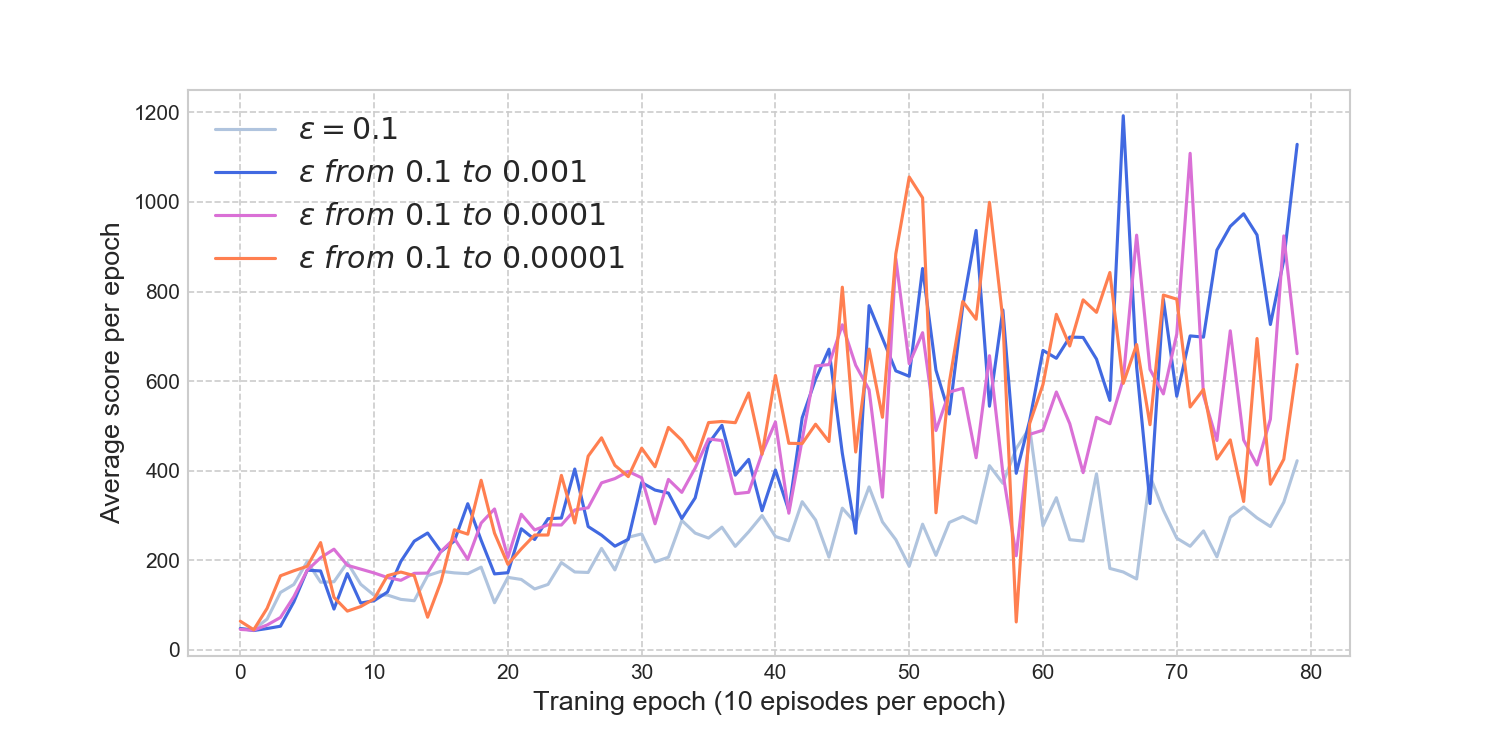}
\caption{Hyper parameter tuning for explore probability $\epsilon$.}
\label{fig:tune-eps}
\end{figure}

All experiments achieve acceptable results in Figure \ref{fig:tune-eps} except the one with fixed $\epsilon=0.1$. In this case, we select $\epsilon$ from $0.1$ to $0.0001$ but either of those three can be chosen according to this graph.  This experiment also demonstrates the positive effect of linear annealing for $\epsilon$.

\subsection{Explore Step}

\begin{figure}[ht]
\centering
\includegraphics[width=13cm]{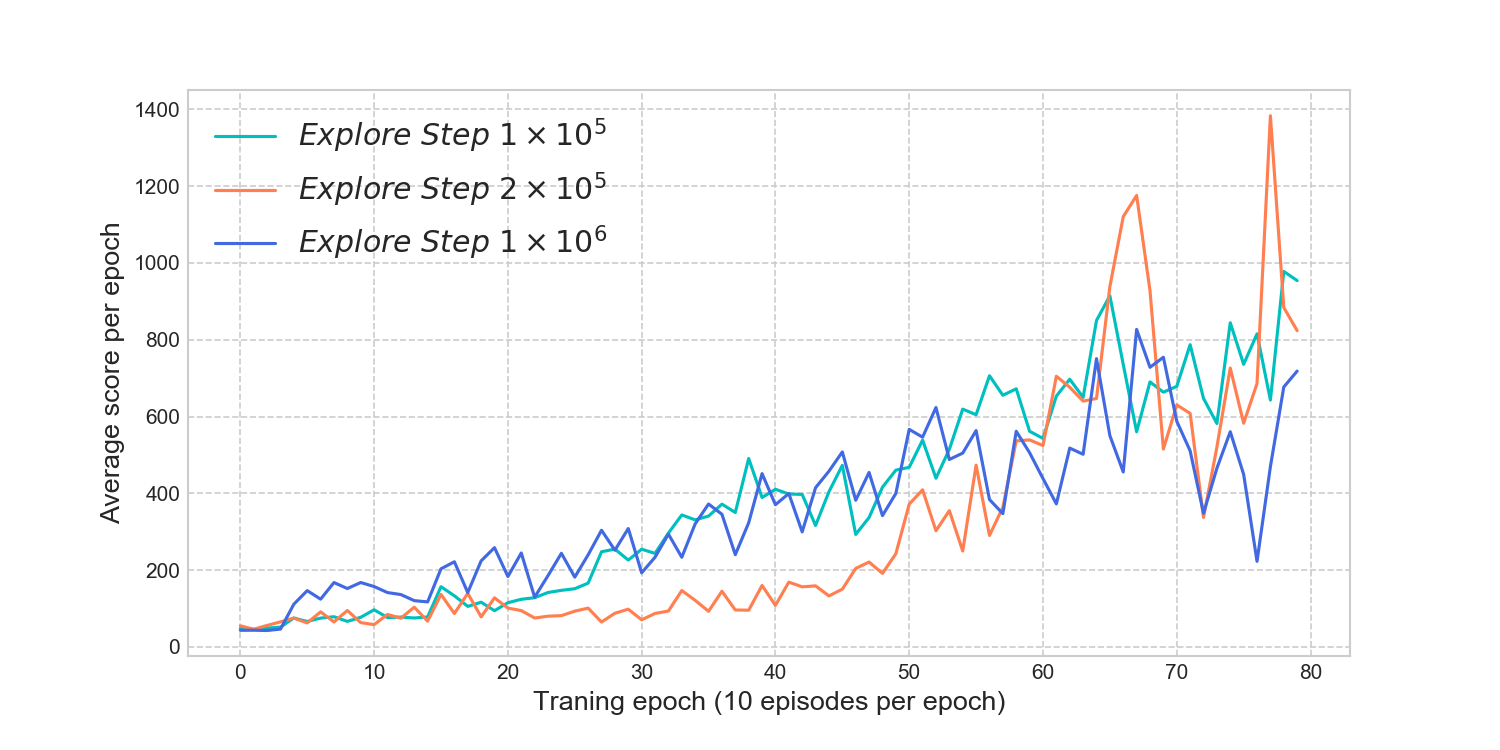}
\caption{Hyper parameter tuning for explore steps.}
\label{fig:tune-step}
\end{figure}

Explore step is the number of steps required to anneal $\epsilon$ from $0.1$ to $0.0001$. As mentioned that hyperparameters related to exploration will not affect too much in this game. The most stable one will be selected from Figure \ref{fig:tune-step} which is $1\times 10^5$.

\subsection{Gamma}

Discount factor decides how far-sighted the agent will be. Too small value will make the agent consider more about the current reward and too big value will make the agent pay the same attention to rewards after this time point. This may confuse the agent about which action leads to a high or low return.

\begin{figure}[ht]
\centering
\includegraphics[width=13cm]{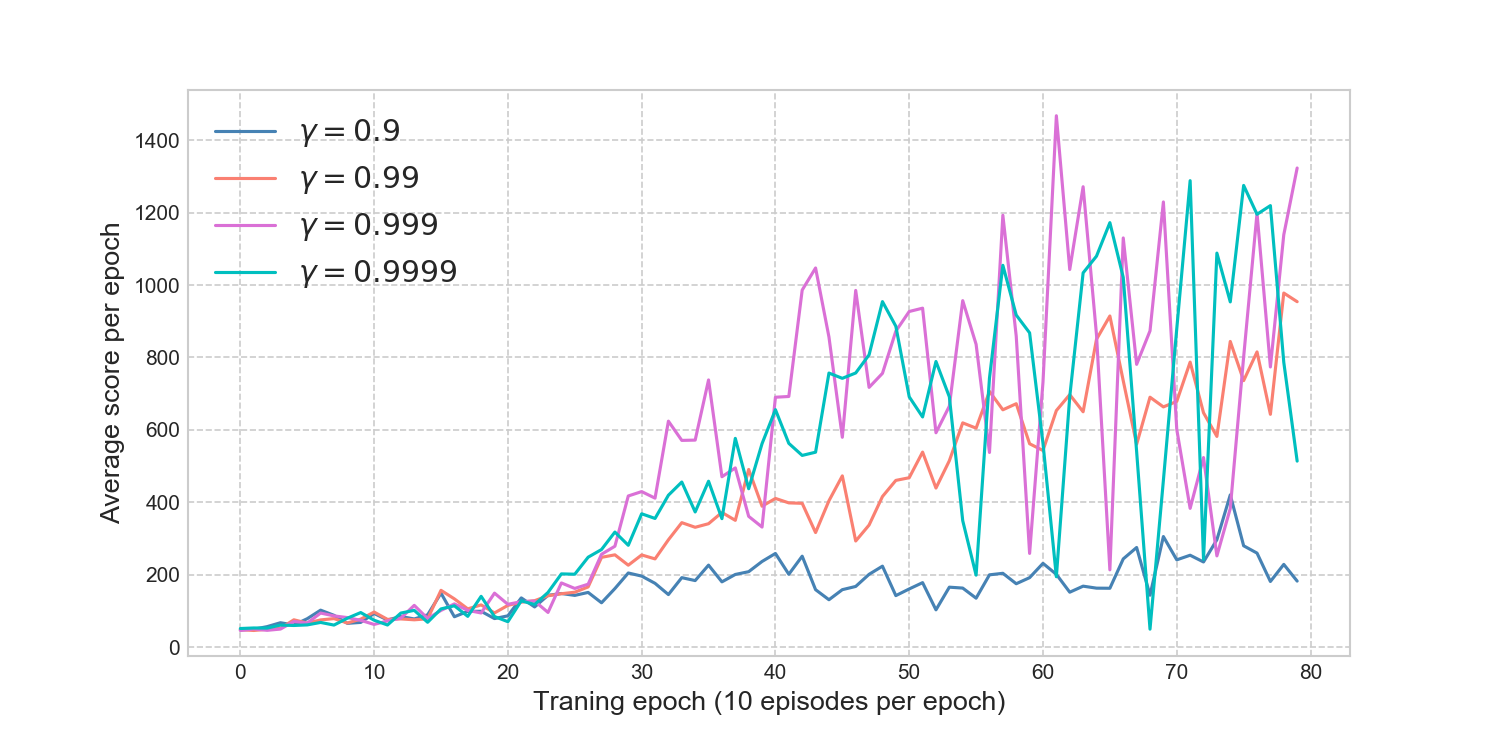}
\caption{Hyper parameter tuning for discount factor $\gamma$.}
\label{fig:tune-gamma}
\end{figure}

Figure \ref{fig:tune-gamma} shows the average score for four different gamma. Obviously, $\gamma=0.9$ make the agent short-sighted and there is no significant change during $80$ epochs. When $\gamma\ge 0.999$, the average score fluctuates widely after 50th epoch. Since $\gamma=0.99$ has a gradually increasing trend, this will be used as the final discount factor.

\section{Training Results}

The tuned hyperparameters from the previous experiment are listed in Table \ref{tab:param-info}. Although these parameters are tuned by DQN algorithm, they are expected to fit other three improved algorithms which are Double DQN, Dueling DQN and DQN with prioritized experience replay because there is no big difference among them. All algorithms will be only trained with 200 epochs because of the time limitation. The total training time for each algorithm is shown in the last column of Table \ref{tab:train-result}

\begin{table}[ht]
\centering
\begin{tabular}{lll}
\hline\hline
\textbf{Hyper parameter} & \textbf{Value before tune} & \textbf{Value after tune} \\
\hline
Memory Size & $3\times 10^5$ & $3\times 10^5$ \\
Batch Size & 128 & 128 \\
Gamma & 0.99 & 0.99 \\
Initial $\epsilon_0$ & $1\times 10^{-1}$ & $1\times 10^{-1}$ \\
Final $\epsilon^\prime$ & $1\times 10^{-3}$ & $1\times 10^{-4}$ \\
Explore steps & $1\times 10^5$ & $1\times 10^5$ \\
Learning Rate & $1\times 10^{-4}$ & $2\times 10^{-5}$ \\
\hline
\end{tabular}
\caption{Hyperparameters used in all experiments}
\label{tab:param-info}
\end{table}

% \begin{table}[ht]
% \centering
% \begin{tabular}{ll}
% \hline\hline
% \textbf{Algorithm} & \textbf{Training time (hours)} \\
% \hline
% DQN & 25.87 \\
% Double DQN & 21.36 \\
% Dueling DQN & 35.78 \\
% DQN with PER & 3.31 \\
% DQN with BN & 32.59 \\
% Double DQN with BN & 29.40 \\
% Dueling DQN with BN & 40.12 \\
% DQN with PER and BN & 3.44 \\
% \hline
% \end{tabular}
% \caption{Training time summary of each algorithm}
% \label{tab:train-time}
% \end{table}

\subsection{DQN}

\begin{figure}[ht]
\centering
\includegraphics[width=13cm]{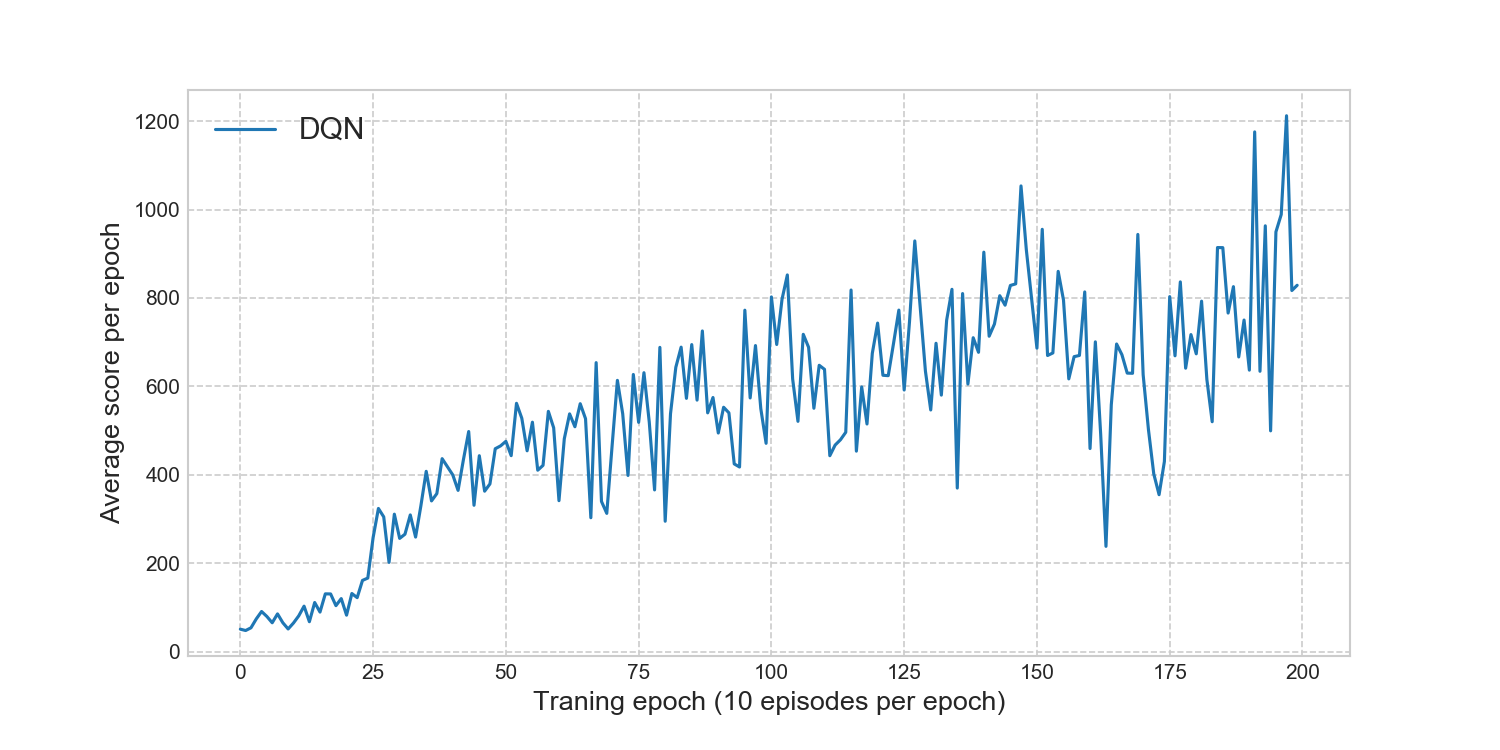}
\caption{Training result for DQN.}
\label{fig:exp-dqn}
\end{figure}

Figure \ref{fig:exp-dqn} shows the result of DQN algorithm for $200$ epochs with tuned parameters. A gradually increased average score can be seen from this graph. This not only proves that the agent can play the game through DQN but also shows that the design of the reward function is relatively reasonable. This result will be treated as a baseline and will be used to compare with other algorithms.

\subsection{Double DQN}

\begin{figure}[ht]
\centering
\includegraphics[width=13cm]{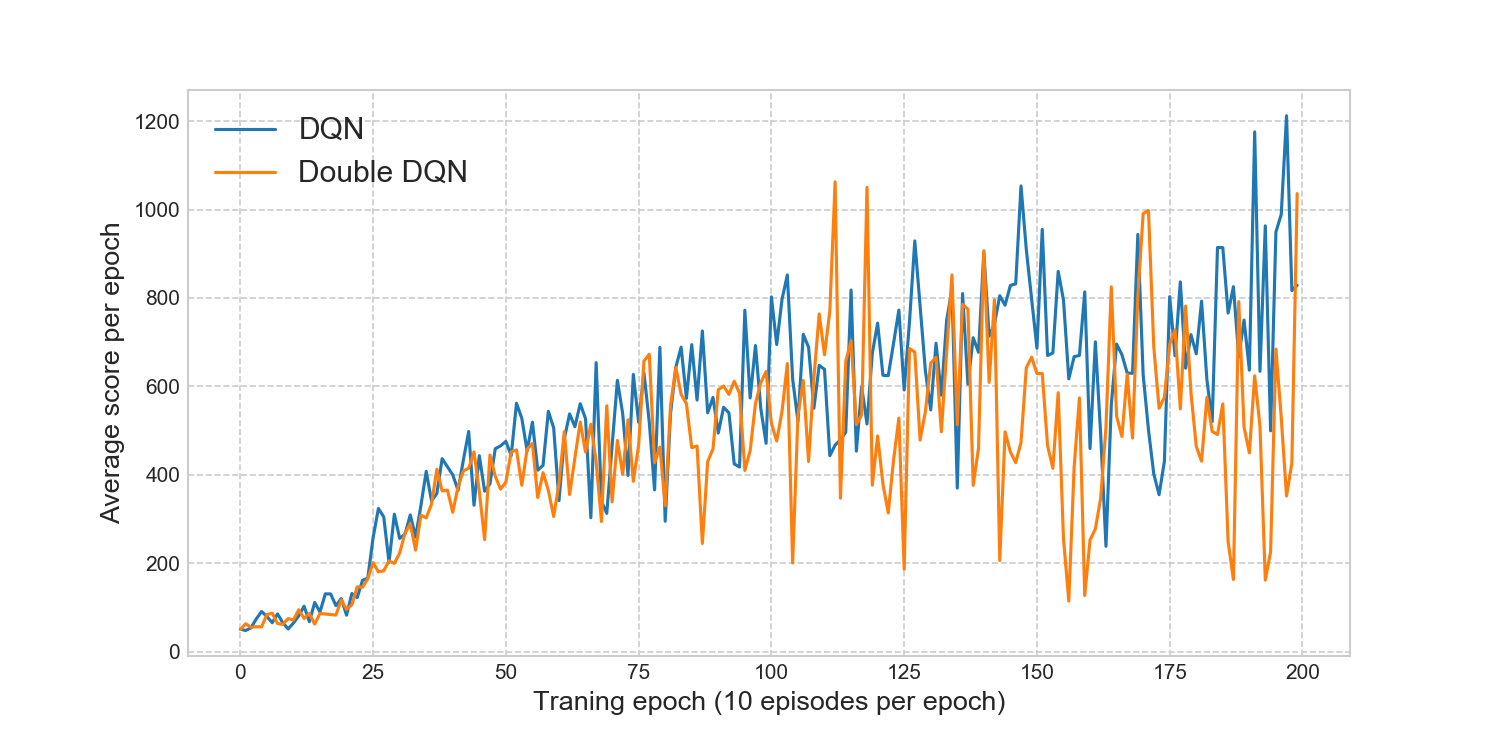}
\caption{Training result for Double DQN compared with DQN.}
\label{fig:exp-double-dqn}
\end{figure}

Double DQN has a similar performance in training compared with DQN. As mentioned before, the effect of $q$ overestimation is not so significant in T-rex Runner because there are only two actions. As shown in Figure \ref{fig:exp-double-dqn}, there are four data points with average scores below $200$ while all average scores are above this value in DQN.

\subsection{Dueling DQN}

\begin{figure}[ht]
\centering
\includegraphics[width=13cm]{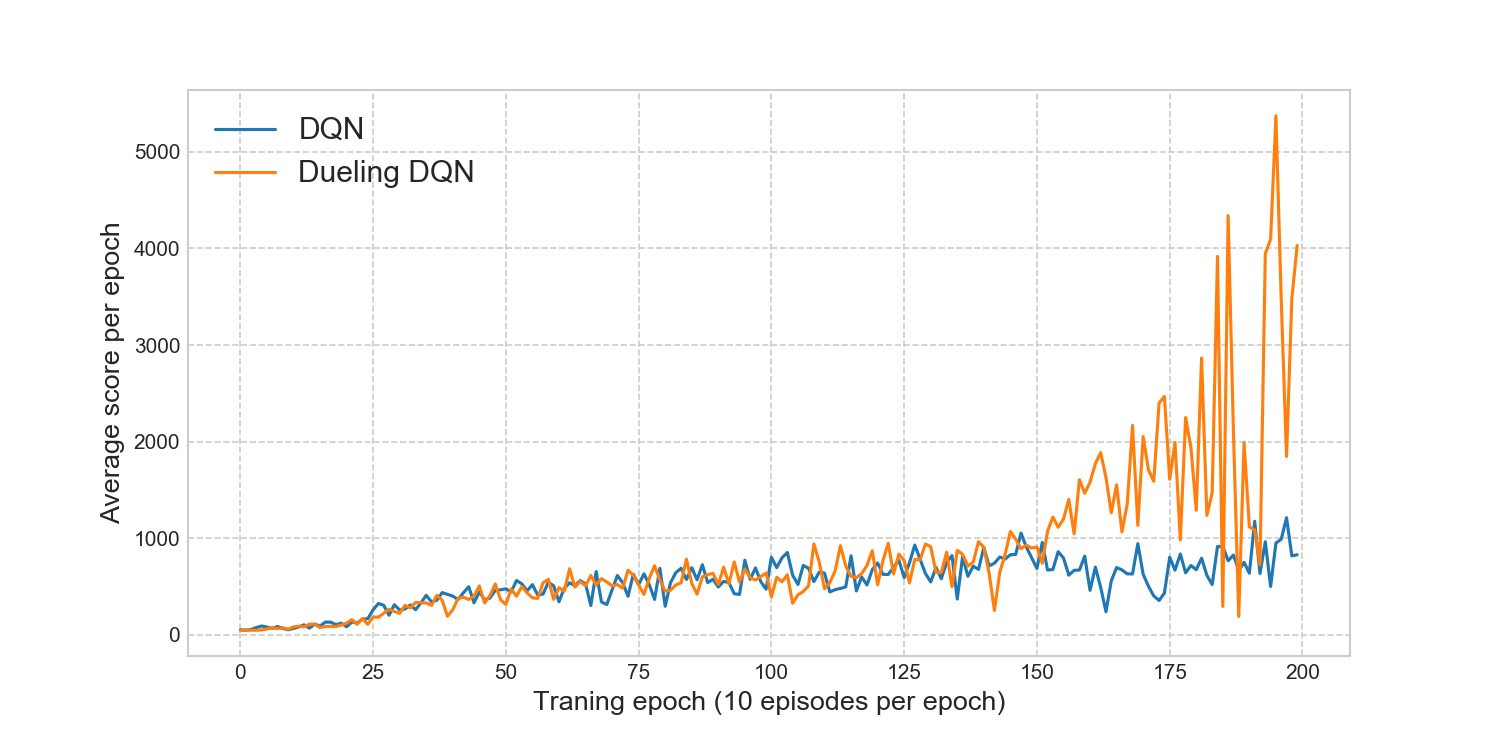}
\caption{Training result for Dueling DQN compared with DQN.}
\label{fig:exp-duel-dqn}
\end{figure}

Surprisingly, dueling DQN shows an incredible training performance after $150$th epoch while the curve before that time seems similar. In Figure \ref{fig:exp-duel-dqn}, the average score is above $5000$ which is ten times higher than the maximum average score in DQN. However, these scores have a high variance which fluctuates widely between $1000$ and $5000$. From the graph, the training process of dueling DQN is stable before $150$th epoch and end up with an increasing trend. Since we tuned all hyperparameters based on DQN, these values may not be the best for dueling network which results in the stable and relatively low average scores before 150th epoch. 

\subsection{DQN with Prioritized Experience Replay}

\begin{figure}[ht]
\centering
\includegraphics[width=13cm]{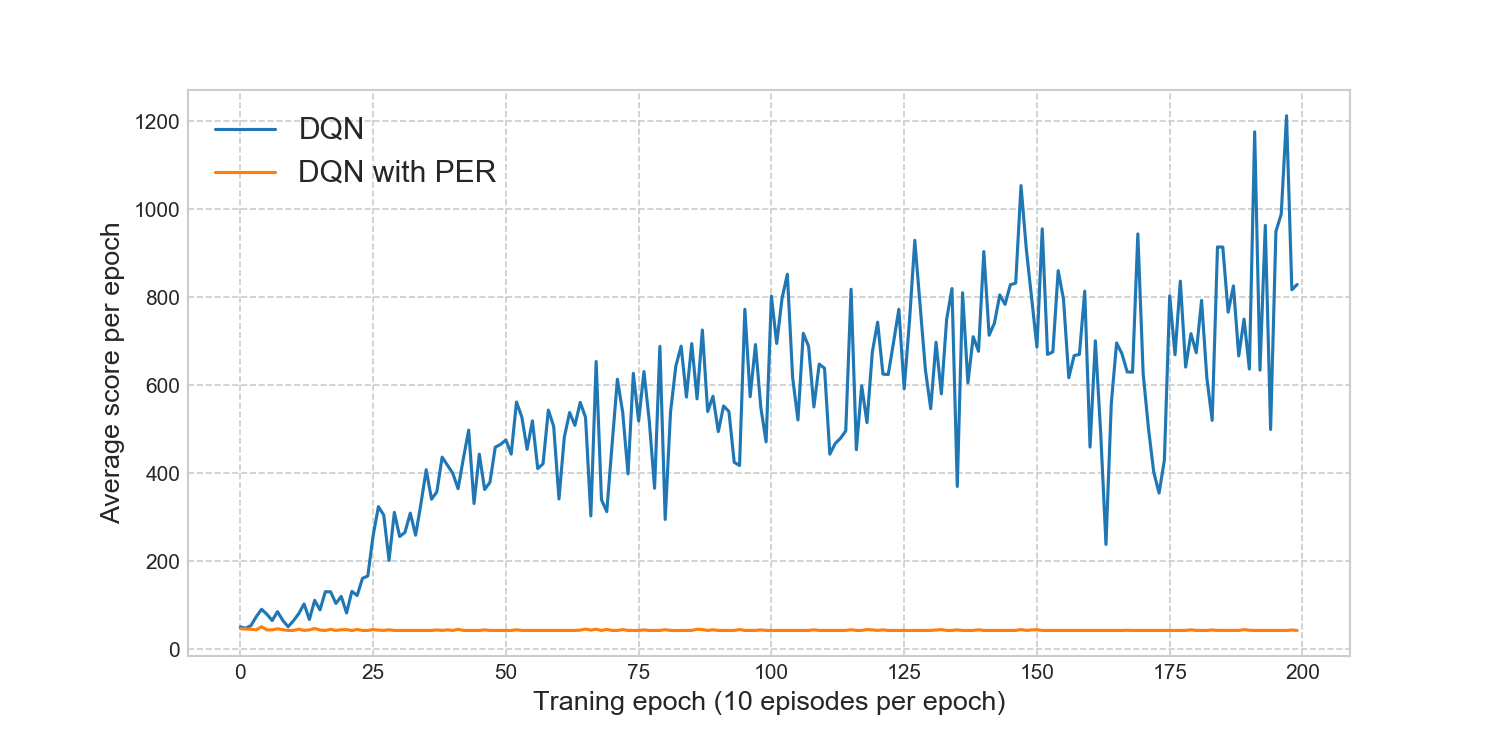}
\caption{Training result for DQN with prioritized experience replay compared with DQN.}
\label{fig:exp-dqn-per}
\end{figure}

Another important finding in this section is the performance of prioritized experience replay. This is expected to have a shorter training time and a higher performance compared with DQN. But the result shown in Figure \ref{fig:exp-dqn-per} suggests that the agent failed to learn to play the game with this method. There are two reasons for that.

One problem is from the algorithm. Compared with DQN, there are two extra steps have been applied to PER: weight calculation and prioritization update. Following the implementation in \cite{schaul2015prioritized}, sum tree which is a data structure with time complexity $O(\log N)$ for sampling and updating is used to store transitions instead of a linear list to accelerate memory related manipulation. The training time of PER is twice more than the one of DQN because of the batch size. Since we know that all sampled transitions will be traversed when updating the prioritization, the larger batch size is the longer time is required to perform this operation. Table \ref{tab:exp-step-difference-per} shows that this process is very time-consuming even using the batch size $32$. These data are extracted from the training results choosing the same score of 43. The step size is the average value from ten records. 

\begin{table}[ht]
\centering
\begin{tabular}{llll}
\hline\hline
\textbf{Algorithm} & \textbf{Score} & \textbf{Batch Size} & \textbf{Step Size} \\
\hline
DQN & 43 & 128 & 180 \\
DQN with PER & 43 & 128 & 7 \\
DQN with PER & 43 & 32 & 22 \\
\hline
\end{tabular}
\caption{Step size difference between DQN and DQN with PER}
\label{tab:exp-step-difference-per}
\end{table}

The other problem is from the game. Because this game is based on Chrome, it continues running when performing optimization while the game from official OpenAI Gym is paused during this operation. Therefore, there is a delayed time before sending the action to Chrome. This influence is enlarged in prioritized experience replay since the time for update operation with batch size $128$ takes approximately $10$ times longer than normal DQN.

Change the choice of hyperparameter can mitigate the first problem but the result is not as good as other algorithms. One thing we can expect is PER is unable to help the agent to get a higher score under this circumstance because the game speed will increase as time goes by. Since the time for updating the prioritization will not change too much, the time interval between two consistent decisions will be longer. This may limit the performance of the model. To eliminate the high computational effect from updating prioritization, the best way is to redevelop the game but due to the time limitation and the primary objective of this study, this result will be used as we can still compare the effect of batch normalization on this algorithm.

\subsection{Batch Normalization}

\begin{figure}[ht]
\centering
\includegraphics[width=15cm]{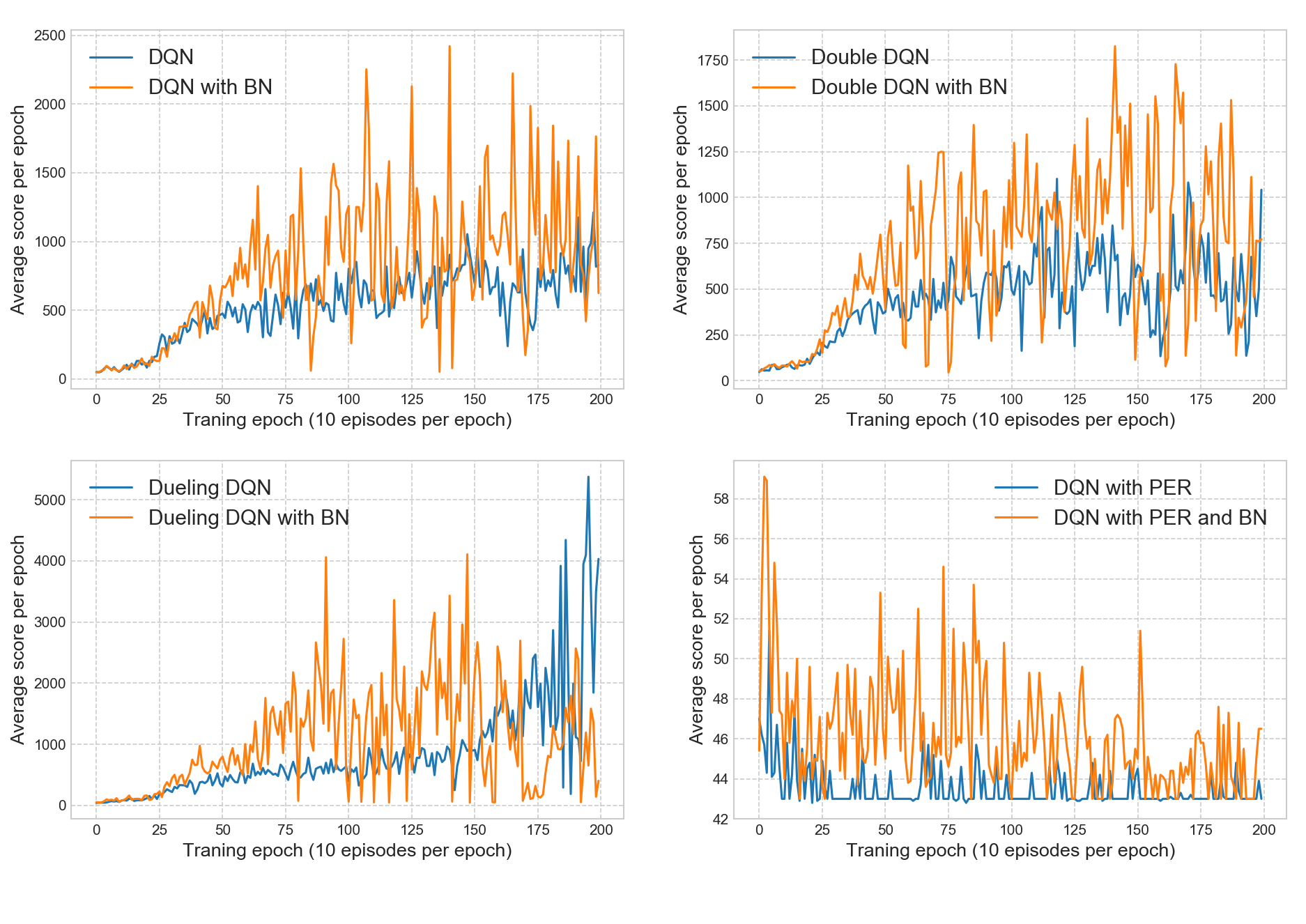}
\caption{Batch normalization on DQN, Double DQN, Dueling DQN and DQN with prioritized experience replay}
\label{fig:exp-bn}
\end{figure}

Since the aim of this experiment is to find how batch normalization affects DQN algorithms, each result will be compared with the one without batch normalization which is shown in Figure \ref{fig:exp-bn}.

From Figure \ref{fig:exp-bn}, we can see that batch normalization can increase the mean of average scores in all experiments. But this also brings high variance which makes the average score diverge. According to the top-left graph, the first time for DQN agent to reach the average $1000$ is approximately $150$th epoch while the agent using DQN with batch normalization reach the same average score at $60$th epoch and it is easy for it to get the higher score after that time. Double DQN curve has a similar trend but batch normalization in both of them also result in wide fluctuation. It is hard to say whether dueling network benefits from the batch normalization because there is a significant increase trend on the bottom left graph. However, it is still can be seen that BN enable the agent to reach the same performance much earlier from $20$th epoch to $90$th epoch. For DQN with prioritized experience replay, even the performance is limited by the game itself, the one with batch normalization still can get a relatively higher score.

\subsection{Further Discussion}

As graphical results and some explanation of them are shown above, this part will discuss numerical results from the experiments. Table \ref{tab:train-result} shows some statistical data fro training process. The maximum score is pointless in most games but considering T-rex Runner is a racing game, we still include this in the table. The last three columns are percentile data which are calculated by sorting in ascending order and finding the $x\%$ observation. So $50\%$ is the same as the median. The last column shows the training time for each algorithm.

\begin{table}[ht]
\centering
\begin{tabular}{llllllll}
\hline\hline
\textbf{Algorithm} & \textbf{Mean} & \textbf{Std} & \textbf{Max}& \textbf{25\%}& \textbf{50\%}& \textbf{75\%}&\textbf{Time (h)}\\
\hline
DQN                 & 537.50 & 393.61 & 1915 & 195.75 & 481 & 820  & 25.87\\
Double DQN          & 443.31 & 394.01 & 2366 & 97.75  & 337 & 662.25   & 21.36\\
Dueling DQN         & 839.04 & 1521.40 & 25706 & 155 & 457 & 956.5  & 35.78\\
DQN with PER        & 43.50  & 2.791 & 71 & 43 & 43  & 43  & 3.31\\
DQN (BN)         & 777.54 & 917.26 & 8978 & 97.75 & 462.5 & 1139.25  & 32.59\\
Double DQN (BN)  & 696.43 & 758.81 & 5521 & 79 & 430.5 & 1104.25 & 29.40 \\
Dueling DQN (BN) & 1050.26 & 1477.00 & 14154 & 84 & 541.5 & 1520  & 40.12\\
DQN with PER (BN) & 46.14 & 7.54 & 98 & 43 & 43 & 43  & 3.44\\  
\hline
\end{tabular}
\caption{Training results}
\label{tab:train-result}
\end{table}

Ignoring the result from prioritized experience replay because of the inappropriate game environment, all algorithms achieve great results according to Table \ref{tab:train-result}. Two algorithms with dueling network stand out from them. The one with batch normalization has the mean over $1000$ which is $200$ more than the one without BN. But the later one got the maximum score of $25706$ which means the agent can keep running for around half an hour in one episode. However, both of them have high variance which exceed the mean.

Double DQN both with BN and without BN perform worse than DQN. This indicates that double DQN may reduce the performance in low dimension action space. But batch normalization shortens the gap between those two algorithms which can be seen from the median and $75\%$ percentile. 

Although most of statistical metrics are improved by batch normalization, the variance is much higher than before. As shown in the table, the variance from DQN with BN is twice more than the one without BN. Only the variance from dueling network is lower after BN. But it is reasonable because there is an incredible increase in the very later stage of the training shown in Figure \ref{fig:exp-bn}. 

\section{Testing Results}

\begin{figure}[ht]
\centering
\includegraphics[width=15cm]{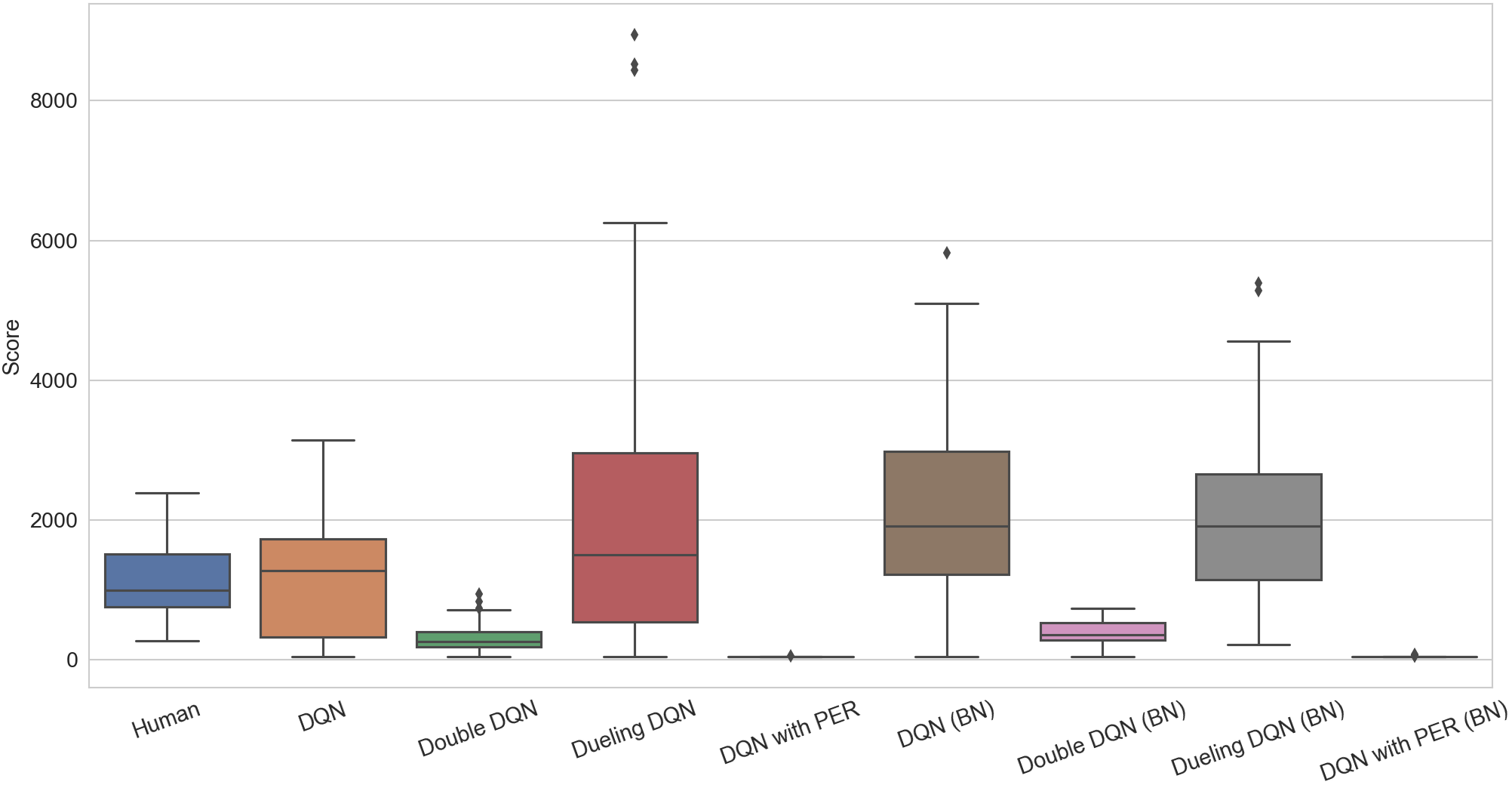}
\caption{Boxplot for test result with eight different algorithms}
\label{fig:eva-results}
\end{figure}

After training the agent for $2000$ episodes, we use the latest model with greedy policy and play T-rex Runner for $30$ times with each algorithm. Figure \ref{fig:eva-results} shows the boxplot of those results as well as the collected data from the human expert. It is obvious that the agent trained by DQN with prioritized experience replay fail to learn to play the game because of the game environment issue discussed in the last section. It is surprising that the performance of double DQN is far from satisfactory even though it has similar training results compared with DQN. Table \ref{tab:test-result} shows that the mean of DQN results is three times higher than the one from double DQN. Dueling DQN algorithm achieves the highest score even though it still has the highest variance which is three times more than the variance from DQN.

According to Table \ref{tab:test-result}, batch normalization improves the performance of the model regardless of algorithms and even the mean of DQN with PER is increased. However, it is not easy to say the effect of BN in dueling DQN is positive or not. From Figure \ref{fig:eva-results}, the one without BN has more outliers which results in high variance even though its mean is higher. Consider the median which is not sensitive with the outlier data, the one with BN is better and the minimum score is more than 200 which stands out from other algorithms. Since score 43 indicates the first time the agent meets the obstacle, it is easy to infer that all trained model fails to jump over the first cacti at least once except dueling DQN with BN. But dueling DQN is not fully trained which can be seen from the training result in Figure \ref{fig:exp-duel-dqn}. That's also one reason for high variance as we can see in the boxplot. The agent trained with dueling DQN achieved over 8000 at least three times. 

\begin{table}[ht]
\centering
\begin{tabular}{llllllll}
\hline\hline
\textbf{Algorithm} & \textbf{Mean} & \textbf{Std} & \textbf{Min}& \textbf{Max}& \textbf{25\%}& \textbf{50\%}& \textbf{75\%}\\
\hline
\textbf{Human} & \textbf{1121.9} & \textbf{499.91} & \textbf{268} & \textbf{2384} & \textbf{758} & \textbf{992.5} & \textbf{1508.5} \\
DQN                 & 1161.30 & 814.36 & 45  & 3142 & 321.5   & 1277   & 1729.5  \\
Double DQN          & 340.93 & 251.40 & 43  & 942  & 178.75  & 259.5  & 400.75  \\
Dueling DQN         & 2383.03 & 2703.64 & 44  & 8943 & 534.75  & 1499.5 & 2961 \\
DQN with PER        & 43.30     & 1.64 & 43  & 52   & 43      & 43     & 43   \\
DQN (BN)         & 2119.47 & 1595.49 & 44  & 5823 & 1218.75 & 1909.5 & 2979.75 \\
Double DQN (BN)  & 382.17 & 188.74 & 43  & 738  & 283.75  & 356    & 525.5   \\
Dueling DQN (BN) & 2083.37 & 1441.50 & 213 & 5389 & 1142.5  & 1912.5 & 2659.75 \\
DQN with PER (BN) & 45.43 & 7.384 & 43  & 78   & 43      & 43     & 43     \\
\hline
\end{tabular}
\caption{Test results}
\label{tab:test-result}
\end{table}

\chapter{Conclusions and Future Works}

The project aims to create an agent trained by four types of algorithms to play T-rex Runner and investigate the influence of batch normalization in reinforcement learning.

The former aim is reached except the prioritized experience replay due to the game environment issue. However, all other algorithms are successfully implemented and achieve great results, especially DQN and dueling DQN. Both of them can achieve better results than human experts. Batch normalization has shown relatively positive effects for all DQN algorithms in this project despite the unstable average score in the training stage.

In further studies, the game environment should be first redeveloped to add a pause function when the neural network is calculating $q$ values or doing optimization. Prioritized experience replay can be tested after that. In this project, only the proportional based method has been implemented, so rank-based prioritization can also be investigated in the future. Further combination of algorithms can be developed such as dueling DQN with prioritized experience replay. Policy-based algorithms such as PPO can also be implemented to train the agent.

There is one interesting idea which has not been implemented yet. Considering the moving speed of the obstacles are gradually increasing, we can divide the game into several stages. Each stage has a neural network which is initialized by the previous stage and will be trained independently. The intuition of this idea is that the consequence of jumping will change when the agent is running in different stages. This may also be one of the reasons for a high variance because when the agent has learned how to get a better score in the later stage, it forgets the best policy in the early stage.

% -------------------------------------------------------------------
% Bibliography
% -------------------------------------------------------------------

\bibliographystyle{abbrv}
\bibliography{mybibliography}

\begin{thebibliography}{10}

\bibitem{allen1971mean}
D.~M. Allen.
\newblock Mean square error of prediction as a criterion for selecting
  variables.
\newblock {\em Technometrics}, 13(3):469--475, 1971.

\bibitem{barto1983neuronlike}
A.~G. Barto, R.~S. Sutton, and C.~W. Anderson.
\newblock Neuronlike adaptive elements that can solve difficult learning
  control problems.
\newblock {\em IEEE transactions on systems, man, and cybernetics},
  (5):834--846, 1983.

\bibitem{bellman1957markov}
R.~Bellman.
\newblock A markov decision process. journal of mathematical mechanics.
\newblock 1957.

\bibitem{bellman1958combinatorial}
R.~Bellman.
\newblock Combinatorial processes and dynamic programming.
\newblock Technical report, RAND CORP SANTA MONICA CA, 1958.

\bibitem{bellman1954theory}
R.~Bellman et~al.
\newblock The theory of dynamic programming.
\newblock {\em Bulletin of the American Mathematical Society}, 60(6):503--515,
  1954.

\bibitem{boureau2010learning}
Y.-L. Boureau, F.~Bach, Y.~LeCun, and J.~Ponce.
\newblock Learning mid-level features for recognition.
\newblock In {\em 2010 IEEE Computer Society Conference on Computer Vision and
  Pattern Recognition}, pages 2559--2566. Citeseer, 2010.

\bibitem{brockman2016openai}
G.~Brockman, V.~Cheung, L.~Pettersson, J.~Schneider, J.~Schulman, J.~Tang, and
  W.~Zaremba.
\newblock Openai gym.
\newblock {\em arXiv preprint arXiv:1606.01540}, 2016.

\bibitem{clark1955generalization}
W.~A. Clark and B.~G. Farley.
\newblock Generalization of pattern recognition in a self-organizing system.
\newblock In {\em Proceedings of the March 1-3, 1955, western joint computer
  conference}, pages 86--91. ACM, 1955.

\bibitem{deng2009imagenet}
J.~Deng, W.~Dong, R.~Socher, L.-J. Li, K.~Li, and L.~Fei-Fei.
\newblock Imagenet: A large-scale hierarchical image database.
\newblock In {\em 2009 IEEE conference on computer vision and pattern
  recognition}, pages 248--255. Ieee, 2009.

\bibitem{fukushima1980neocognitron}
K.~Fukushima.
\newblock Neocognitron: A self-organizing neural network model for a mechanism
  of pattern recognition unaffected by shift in position.
\newblock {\em Biological cybernetics}, 36(4):193--202, 1980.

\bibitem{ge2015escaping}
R.~Ge, F.~Huang, C.~Jin, and Y.~Yuan.
\newblock Escaping from saddle points—online stochastic gradient for tensor
  decomposition.
\newblock In {\em Conference on Learning Theory}, pages 797--842, 2015.

\bibitem{he2015resnet}
K.~He, X.~Zhang, S.~Ren, and J.~Sun.
\newblock Resnet-deep residual learning for image recognition.
\newblock {\em ResNet: Deep Residual Learning for Image Recognition}, 2015.

\bibitem{hu2012matrix}
P.~Hu.
\newblock Matrix calculus: Derivation and simple application.
\newblock Technical report, 2012.

\bibitem{huang2017densely}
G.~Huang, Z.~Liu, L.~Van Der~Maaten, and K.~Q. Weinberger.
\newblock Densely connected convolutional networks.
\newblock In {\em Proceedings of the IEEE conference on computer vision and
  pattern recognition}, pages 4700--4708, 2017.

\bibitem{ioffe2015batch}
S.~Ioffe and C.~Szegedy.
\newblock Batch normalization: Accelerating deep network training by reducing
  internal covariate shift.
\newblock {\em arXiv preprint arXiv:1502.03167}, 2015.

\bibitem{klopf1972brain}
A.~H. Klopf.
\newblock Brain function and adaptive systems: a heterostatic theory.
\newblock Technical report, AIR FORCE CAMBRIDGE RESEARCH LABS HANSCOM AFB MA,
  1972.

\bibitem{klopf1982hedonistic}
A.~H. Klopf.
\newblock {\em The hedonistic neuron: a theory of memory, learning, and
  intelligence}.
\newblock Toxicology-Sci, 1982.

\bibitem{konda2000actor}
V.~R. Konda and J.~N. Tsitsiklis.
\newblock Actor-critic algorithms.
\newblock In {\em Advances in neural information processing systems}, pages
  1008--1014, 2000.

\bibitem{krizhevsky2012imagenet}
A.~Krizhevsky, I.~Sutskever, and G.~E. Hinton.
\newblock Imagenet classification with deep convolutional neural networks.
\newblock In {\em Advances in neural information processing systems}, pages
  1097--1105, 2012.

\bibitem{lawrence1997face}
S.~Lawrence, C.~L. Giles, A.~C. Tsoi, and A.~D. Back.
\newblock Face recognition: A convolutional neural-network approach.
\newblock {\em IEEE transactions on neural networks}, 8(1):98--113, 1997.

\bibitem{lecun1998gradient}
Y.~LeCun, L.~Bottou, Y.~Bengio, P.~Haffner, et~al.
\newblock Gradient-based learning applied to document recognition.
\newblock {\em Proceedings of the IEEE}, 86(11):2278--2324, 1998.

\bibitem{lin1992self}
L.-J. Lin.
\newblock Self-improving reactive agents based on reinforcement learning,
  planning and teaching.
\newblock {\em Machine learning}, 8(3-4):293--321, 1992.

\bibitem{lin2013network}
M.~Lin, Q.~Chen, and S.~Yan.
\newblock Network in network.
\newblock {\em arXiv preprint arXiv:1312.4400}, 2013.

\bibitem{mcculloch1943logical}
W.~S. McCulloch and W.~Pitts.
\newblock A logical calculus of the ideas immanent in nervous activity.
\newblock {\em The bulletin of mathematical biophysics}, 5(4):115--133, 1943.

\bibitem{minsky1969perceptron}
M.~Minsky and S.~Papert.
\newblock Perceptron: an introduction to computational geometry.
\newblock {\em The MIT Press, Cambridge, expanded edition}, 19(88):2, 1969.

\bibitem{mnih2016asynchronous}
V.~Mnih, A.~P. Badia, M.~Mirza, A.~Graves, T.~Lillicrap, T.~Harley, D.~Silver,
  and K.~Kavukcuoglu.
\newblock Asynchronous methods for deep reinforcement learning.
\newblock In {\em International conference on machine learning}, pages
  1928--1937, 2016.

\bibitem{mnih2013playing}
V.~Mnih, K.~Kavukcuoglu, D.~Silver, A.~Graves, I.~Antonoglou, D.~Wierstra, and
  M.~Riedmiller.
\newblock Playing atari with deep reinforcement learning.
\newblock {\em arXiv preprint arXiv:1312.5602}, 2013.

\bibitem{mnih2015human}
V.~Mnih, K.~Kavukcuoglu, D.~Silver, A.~A. Rusu, J.~Veness, M.~G. Bellemare,
  A.~Graves, M.~Riedmiller, A.~K. Fidjeland, G.~Ostrovski, et~al.
\newblock Human-level control through deep reinforcement learning.
\newblock {\em Nature}, 518(7540):529, 2015.

\bibitem{neal2001annealed}
R.~M. Neal.
\newblock Annealed importance sampling.
\newblock {\em Statistics and computing}, 11(2):125--139, 2001.

\bibitem{ng1999policy}
A.~Y. Ng, D.~Harada, and S.~Russell.
\newblock Policy invariance under reward transformations: Theory and
  application to reward shaping.
\newblock In {\em ICML}, volume~99, pages 278--287, 1999.

\bibitem{rosenblatt1957perceptron}
F.~Rosenblatt.
\newblock {\em The perceptron, a perceiving and recognizing automaton Project
  Para}.
\newblock Cornell Aeronautical Laboratory, 1957.

\bibitem{rosenblatt1960perceptron}
F.~Rosenblatt.
\newblock Perceptron simulation experiments.
\newblock {\em Proceedings of the IRE}, 48(3):301--309, 1960.

\bibitem{rumelhart1988learning}
D.~E. Rumelhart, G.~E. Hinton, R.~J. Williams, et~al.
\newblock Learning representations by back-propagating errors.
\newblock {\em Cognitive modeling}, 5(3):1, 1988.

\bibitem{rummery1994line}
G.~A. Rummery and M.~Niranjan.
\newblock {\em On-line Q-learning using connectionist systems}, volume~37.
\newblock University of Cambridge, Department of Engineering Cambridge,
  England, 1994.

\bibitem{samuel1959aerosol}
A.~J. Samuel.
\newblock Aerosol dispensers and like pressurized packages, Sept.~15 1959.
\newblock US Patent 2,904,229.

\bibitem{schaul2015prioritized}
T.~Schaul, J.~Quan, I.~Antonoglou, and D.~Silver.
\newblock Prioritized experience replay.
\newblock {\em arXiv preprint arXiv:1511.05952}, 2015.

\bibitem{scherer2010evaluation}
D.~Scherer, A.~M{\"u}ller, and S.~Behnke.
\newblock Evaluation of pooling operations in convolutional architectures for
  object recognition.
\newblock In {\em International conference on artificial neural networks},
  pages 92--101. Springer, 2010.

\bibitem{schulman2017proximal}
J.~Schulman, F.~Wolski, P.~Dhariwal, A.~Radford, and O.~Klimov.
\newblock Proximal policy optimization algorithms.
\newblock {\em arXiv preprint arXiv:1707.06347}, 2017.

\bibitem{silver2015notes}
D.~Silver.
\newblock University college london course on reinforcement learning, 2015.

\bibitem{silver2017mastering}
D.~Silver, T.~Hubert, J.~Schrittwieser, I.~Antonoglou, M.~Lai, A.~Guez,
  M.~Lanctot, L.~Sifre, D.~Kumaran, T.~Graepel, et~al.
\newblock Mastering chess and shogi by self-play with a general reinforcement
  learning algorithm.
\newblock {\em arXiv preprint arXiv:1712.01815}, 2017.

\bibitem{simonyan2014very}
K.~Simonyan and A.~Zisserman.
\newblock Very deep convolutional networks for large-scale image recognition.
\newblock {\em arXiv preprint arXiv:1409.1556}, 2014.

\bibitem{sutton1988learning}
R.~S. Sutton.
\newblock Learning to predict by the methods of temporal differences.
\newblock {\em Machine learning}, 3(1):9--44, 1988.

\bibitem{sutton2018reinforcement}
R.~S. Sutton and A.~G. Barto.
\newblock {\em Reinforcement learning: An introduction}.
\newblock MIT press, 2018.

\bibitem{sutton2000policy}
R.~S. Sutton, D.~A. McAllester, S.~P. Singh, and Y.~Mansour.
\newblock Policy gradient methods for reinforcement learning with function
  approximation.
\newblock In {\em Advances in neural information processing systems}, pages
  1057--1063, 2000.

\bibitem{szegedy2017inception}
C.~Szegedy, S.~Ioffe, V.~Vanhoucke, and A.~A. Alemi.
\newblock Inception-v4, inception-resnet and the impact of residual connections
  on learning.
\newblock In {\em Thirty-First AAAI Conference on Artificial Intelligence},
  2017.

\bibitem{szegedy2015going}
C.~Szegedy, W.~Liu, Y.~Jia, P.~Sermanet, S.~Reed, D.~Anguelov, D.~Erhan,
  V.~Vanhoucke, and A.~Rabinovich.
\newblock Going deeper with convolutions.
\newblock In {\em Proceedings of the IEEE conference on computer vision and
  pattern recognition}, pages 1--9, 2015.

\bibitem{thorndike1911animal}
E.~Thorndike.
\newblock Animal intelligence; experimental studies, by edward l. thorndike,
  1911.

\bibitem{tieleman2012lecture}
T.~Tieleman and G.~Hinton.
\newblock Lecture 6.5-rmsprop, coursera: Neural networks for machine learning.
\newblock {\em University of Toronto, Technical Report}, 2012.

\bibitem{vaillant1994original}
R.~Vaillant, C.~Monrocq, and Y.~Le~Cun.
\newblock Original approach for the localisation of objects in images.
\newblock {\em IEE Proceedings-Vision, Image and Signal Processing},
  141(4):245--250, 1994.

\bibitem{van2016deep}
H.~Van~Hasselt, A.~Guez, and D.~Silver.
\newblock Deep reinforcement learning with double q-learning.
\newblock In {\em Thirtieth AAAI conference on artificial intelligence}, 2016.

\bibitem{waibel1995phoneme}
A.~Waibel, T.~Hanazawa, G.~Hinton, K.~Shikano, and K.~J. Lang.
\newblock Phoneme recognition using time-delay neural networks.
\newblock {\em Backpropagation: Theory, Architectures and Applications}, pages
  35--61, 1995.

\bibitem{wang2015dueling}
Z.~Wang, T.~Schaul, M.~Hessel, H.~Van~Hasselt, M.~Lanctot, and N.~De~Freitas.
\newblock Dueling network architectures for deep reinforcement learning.
\newblock {\em arXiv preprint arXiv:1511.06581}, 2015.

\bibitem{watkins1989learning}
C.~J. C.~H. Watkins.
\newblock Learning from delayed rewards.
\newblock 1989.

\bibitem{widrow1973punish}
B.~Widrow, N.~K. Gupta, and S.~Maitra.
\newblock Punish/reward: Learning with a critic in adaptive threshold systems.
\newblock {\em IEEE Transactions on Systems, Man, and Cybernetics},
  (5):455--465, 1973.

\bibitem{widrow1960adaptive}
B.~Widrow and M.~E. Hoff.
\newblock Adaptive switching circuits.
\newblock Technical report, Stanford Univ Ca Stanford Electronics Labs, 1960.

\end{thebibliography}

% -------------------------------------------------------------------
% Appendices
% -------------------------------------------------------------------

% \begin{appendices}
% \input{sections/appendixA.tex}
% \input{sections/appendixB.tex}
% \end{appendices}

\end{document}